\documentclass[10pt,twocolumn,letterpaper]{article}
\usepackage[T1]{fontenc}

\usepackage{cvpr}
\usepackage{times}
\usepackage{epsfig}
\usepackage{graphicx}
\usepackage{amsmath}
\usepackage{amssymb}
\usepackage[sort,nocompress]{cite}
\usepackage{comment}
\usepackage{relsize}
\usepackage{enumitem}
\usepackage{grffile}
\usepackage{subfig}
\usepackage{placeins}
\PassOptionsToPackage{normalem}{ulem}
\usepackage{ulem}
\usepackage{hyphenat}
\usepackage{multirow}
\usepackage{color}
\usepackage{colortbl}
\usepackage{rotating}
\definecolor{lightgray}{gray}{0.8}
\definecolor{verylightgray}{gray}{0.9}
\definecolor{tango_green}{RGB}{78, 154, 6}

\makeatletter
\renewcommand{\paragraph}{%
\@startsection{paragraph}{4}%
{\z@}{0.5ex \@plus 1ex \@minus .2ex}{-0.5em}%
{\normalfont \normalsize \bfseries}%
}
\makeatother

\usepackage[pagebackref=true,breaklinks=true,letterpaper=true,colorlinks,bookmarks=false]{hyperref}

 \cvprfinalcopy 

\ifcvprfinal\fi

\definecolor{chamoisee}{rgb}{0.63, 0.47, 0.35}

\newcommand{\key}{DBLP:conf/iccv/LeeKG11}
\newcommand{\nlc}{Faktor2014Bmvc}

\newcommand{\refsec}[1]{\S\ref{sec:#1}}

\newcommand{\segtrack}{SegTrack-v2}

\newcommand{\dagvos}{DBLP:conf/cvpr/ZhangJS13}

\begin{document}

\title{Learning Video Object Segmentation from Static Images}




\author{
	\textsuperscript{*}Anna Khoreva\textsuperscript{3}\hspace{1.0em}
	\textsuperscript{*}Federico Perazzi\textsuperscript{1,2}\hspace{1.0em}
	Rodrigo Benenson\textsuperscript{3}\hspace{1.0em} \\
	Bernt Schiele\textsuperscript{3}\hspace{1.0em}
	Alexander Sorkine-Hornung\textsuperscript{1}
	\\\\
	\textsuperscript{1}Disney Research \hspace{1.0em} \textsuperscript{2}ETH Zurich\\
	\textsuperscript{3}Max Planck Institute for Informatics, Saarbr{\"u}cken, Germany\\
}

\maketitle
\vspace{-0.5em}

\begin{abstract}

Inspired by recent advances of deep learning in instance segmentation and object tracking, we introduce video object segmentation problem as a concept of guided instance segmentation.
Our model proceeds on a per-frame basis, guided by the output of the previous frame towards the object of interest in the next frame.
We demonstrate that highly accurate object segmentation in videos can be enabled by using a convnet trained with static images only.
The key ingredient of our approach is a combination of offline and online learning strategies, where the former serves
to produce a refined mask from the previous' frame estimate and the latter allows to capture the appearance of the specific object instance.
Our method can handle different types of input annotations: bounding boxes and segments, as well as
incorporate multiple annotated frames, making the system suitable for diverse applications.
We obtain competitive results on three different datasets, independently from the type of input annotation.
\end{abstract}
\renewcommand{\thefootnote}{\fnsymbol{footnote}}
\footnotetext[1]{The first two authors contributed equally}

\section{Introduction}
\label{sec:introduction}
Convolutional neural networks (convnets) have shown outstanding performance in many fundamental areas in computer vision, enabled by the availability of large-scale annotated datasets
(e.g., ImageNet classification~\cite{Krizhevsky2012Nips,Russakovsky2015Ijcv}).
However, some important challenges in video processing can be difficult to approach using convnets,
since creating a sufficiently large body of densely, pixel-wise annotated video data for training is usually prohibitive.

One example domain is video object segmentation. Given only one or a few frames with segmentation mask annotations of a particular object instance,
the task is to accurately segment the same instance in all other frames of the video.
Current top performing approaches either interleave box tracking and segmentation \cite{Xiao2016Cvpr},
or propagate the first frame segment annotation in space-time via CRF-like and GrabCut-like techniques \cite{Tsai2016Cvpr,Maerki2016Cvpr}.
%

One of the key insights and contributions of this paper is that fully annotated video data is not necessary.
We demonstrate that highly accurate video object segmentation can be enabled using a convnet trained with {\em static images} only.

We approach video object segmentation from a new angle.
We show that a convnet designed for semantic image segmentation~\cite{Chen2016ArxivDeeplabv2} can be utilized to perform per-frame instance segmentation,
i.e., segmentation of generic objects while distinguishing different instances of the same class.
For each new video frame the network is guided towards the object of interest by feeding in the previous' frame mask estimate.
We therefore refer to our approach as \emph{guided instance segmentation}. To the best of our knowledge, it represents the first fully trained approach to video object segmentation.


Our system is efficient due to its feed-forward architecture and can generate high quality results in a single pass over the video, without the need for considering more than one frame at a time.
This is in stark contrast to many other video segmentation approaches,
which usually require global connections over multiple frames or even the whole video sequence in order to achieve coherent results.
The method can handle different types of
annotations and in the extreme case, even simple bounding boxes as input are sufficient,
achieving competitive results, rendering our method flexible with respect to various practical applications.

Key to the video segmentation quality of our approach is a combined offline / online learning strategy.
In the offline phase, we use deformation and coarsening on the image masks in order to train the network to produce accurate output masks from their rough estimates.
An online training phase extends ideas from previous works on object tracking \cite{Danelljan2016Eccv,Nam2016Cvpr} to video segmentation and enables the method to be easily optimized with respect to an object of interest
in a novel input video.

The result is a single, generic system that compares favourably to most classical approaches on three extremely heterogeneous video segmentation benchmarks, despite using the same model and parameters across all videos.
We provide a detailed ablation study and explore the impact of varying number and types of annotations, and moreover discuss extensions of the proposed model
, allowing to improve the quality even further.

\section{Related work}
\label{sec:related-work}

The idea of performing video object segmentation via tracking at the pixel level is at least
a decade old \cite{Ren2007Cvpr}. Recent approaches interweave box
tracking with box-driven segmentation (e.g. $\mathtt{TRS}$ \cite{Xiao2016Cvpr}),
or propagate the first frame segmentation via graph labeling approaches.

\paragraph{Local propagation}
$\mathtt{JOTS}$ \cite{Wen2015Cvpr} builds a graph over neigbhour
frames connecting superpixels and (generic) object parts to solve the video labeling task. $\mathtt{ObjFlow}$
\cite{Tsai2016Cvpr} builds a graph over pixels and superpixels, uses
convnet based appearance terms, and interleaves labeling with
optical flow estimation. Instead of using superpixels or proposals, $\mathtt{BVS}$ \cite{Maerki2016Cvpr}
formulates a fully-connected pixel-level graph between frames and efficiently infer the labeling over the vertices of a spatio-temporal bilateral grid \cite{Chen2007}.
Because these methods propagate information only across neighbor frames
they have difficulties capturing long range relationships and ensuring globally consistent segmentations.

\paragraph{Global propagation}
In order to overcome these limitations, some methods have proposed to use long-range connections between video frames \cite{\key,\nlc,\dagvos}.
In particular, we compare to  $\mathtt{FCP}$ \cite{Perazzi2015Iccv},
$\mathtt{Z15}$ \cite{Zhang2015CvprObjectSegmentation} and $\mathtt{W16}$ \cite{Wang2016Accv} which build a global graph structure over object proposal segments, and then infer a consistent segmentation.
A limitation of methods utilizing long-range connections is that they have to operate on larger image regions such as superpixels or object proposals for acceptable speed and memory usage,
compromising on their ability to handle fine image detail.

\paragraph{Unsupervised segmentation}
Another family of work does general moving object segmentation (over all parts of the image), and selects post-hoc the space-time tube that best match the annotation,
e.g. $\mathtt{NLC}$ \cite{Faktor2014Bmvc} and \cite{Grundmann2010Cvpr,Li2013Iccv,Xiao2016Cvpr}.


In contrast, our approach sides-steps the use of any intermediate tracked box, superpixels, or object proposals and proceeds on a per-frame basis, therefore  efficiently handling even long sequences at full detail.
We focus on propagating the first frame segmentation forward onto future frames, using an online fine-tuned convnet as appearance model for segmenting the object of interest in the next frames.

\paragraph{Box tracking} Some previous works have investigated approaches that improve segmentation quality by leveraging object tracking and
vice versa \cite{Ren2007Cvpr,DBLP:conf/iccv/DuffnerG13,DBLP:conf/iccv/ChockalingamPB09,Xiao2016Cvpr}. More recent, state-of-the-art tracking
methods are based on discriminative correlation filters
over handcrafted features (e.g. HOG) and over frozen deep learned
features \cite{Danelljan2015Iccvw,Danelljan2016Eccv}, or are convnet
based trackers on their own right \cite{Held2016Eccv,Nam2016Cvpr}.

Our approach is most closely related to the latter group.
$\mathtt{GOTURN}$ \cite{Held2016Eccv} proposes
to train offline a convnet so as to directly regress the bounding
box in the current frame based on
the object position and appearance in the previous frame.
$\mathtt{MDNet}$ \cite{Nam2016Cvpr}
proposes to use online fine-tuning of a convnet to model the object appearance.

Our training strategy is inspired by $\mathtt{GOTURN}$ for the offline part,
and $\mathtt{MDNet}$ for the online stage.
Compared to the aforementioned
methods our approach operates at pixel level masks instead of boxes.
Differently from $\mathtt{MDNet}$, we do not replace the domain-specific layers,
instead finetuning all the layers on the available annotations for each individual video sequence.

\paragraph{Instance segmentation}

At each frame, video object segmentation outputs a single instance segmentation.
Given an estimate of the object location and size, bottom-up
segment proposals \cite{PontTuset2016PamiMCG} or GrabCut \cite{Rother2004TogGrabcut}
variants can be used as shape guesses. Also specific convnet architectures
have been proposed for instance segmentation \cite{Hariharan2015Cvpr,Xu2016Cvpr,Pinheiro2015Nips,Pinheiro2016Eccv}.
Our approach outputs per-frame instance segmentation using a convnet architecture, inspired by works
from other domains like \cite{Carreira2016Cvpr,Xu2016Cvpr,Shen2016Cgf}.
%
%
A concurrent work \cite{Caelles2016} also exploits convnets for video object segmentation. Differently from our approach their segmentation is not guided, which might result in performance decay over time. Furthermore, the offline training exploits full video sequence annotations that are notoriously difficult to obtain.

\paragraph{Interactive video segmentation}
Applications such as video editing for movie production often require a level of accuracy beyond the current state-of-the-art.
Thus several works have also considered video segmentation with variable annotation effort, enabling human interaction using clicks \cite{Jain2016Hcomp,SpinaF16,Wang2014Cviu},
or strokes \cite{Bai2009Tog,Zhong2012Tog,Fan2015SiggraphAsia}. In this work we consider instead box or (full) segment annotations on multiple frames.
In \refsec{results} we report results when varying the amount of annotation effort (from one frame per video to all frames).

\section{MaskTrack method}
\label{sec:method}

We approach the video object segmentation problem from a new angle we refer as \emph{guided instance segmentation}.
For each new frame we wish to label pixels as object/non-object of interest, for this we build upon the architecture of the existing pixel labelling convnet and
train it to generate per-frame instance segments.
We pick DeepLabv2 \cite{Chen2016ArxivDeeplabv2},
but our approach is agnostic of the specific architecture selected.

The challenge is then: how to inform the network which instance to segment?
We solve this by using two complementary strategies. One is guiding the network towards the instance of interest by feeding in the previous' frame mask estimate during offline training (\S\ref{sec:method-offline}).
And a second is employing online training (\S\ref{sec:method-online}) to fine-tune the model to become more specialized for the specific instance.


\subsection{Learning to segment instances offline}
\label{sec:method-offline}

In order to guide the pixel labeling network to segment the object of interest, we begin by expanding the convnet input from RGB to RGB+mask channel (4 channels).
The extra mask channel is meant to provide an estimate of the visible area of the object in the current frame, its approximate location and shape.
We can then train the labelling convnet to provide as output an accurate segmentation of the object, given as input the current image and a rough estimate of the object mask.
Our tracking network is de-facto a "mask refinement" network.

There are two key observations that make this approach practical. First, very rough input masks are enough for our trained network to provide sensible output segments.
Even a large bounding box as input will result in a reasonable output (see \S\ref{sec:ablation-study}). The input mask's main role is to point the convnet towards the correct object instance to segment.

Second, this particular approach does not require us to use video as training data, such as done in \cite{Held2016Eccv, Bertinetto2016Arxiv, Nam2016Cvpr}.
Because we only use a mask as additional input, instead of an image crop as in \cite{Held2016Eccv, Bertinetto2016Arxiv}, we can easily synthesize training samples from single frame instance segmentation annotations.
This allows to train from a large set of diverse images and avoids having to use existing (scarce and small) video segmentation benchmarks for training.

\begin{figure}
\begin{centering}
\includegraphics[width=0.9\columnwidth,height=0.17\textheight]{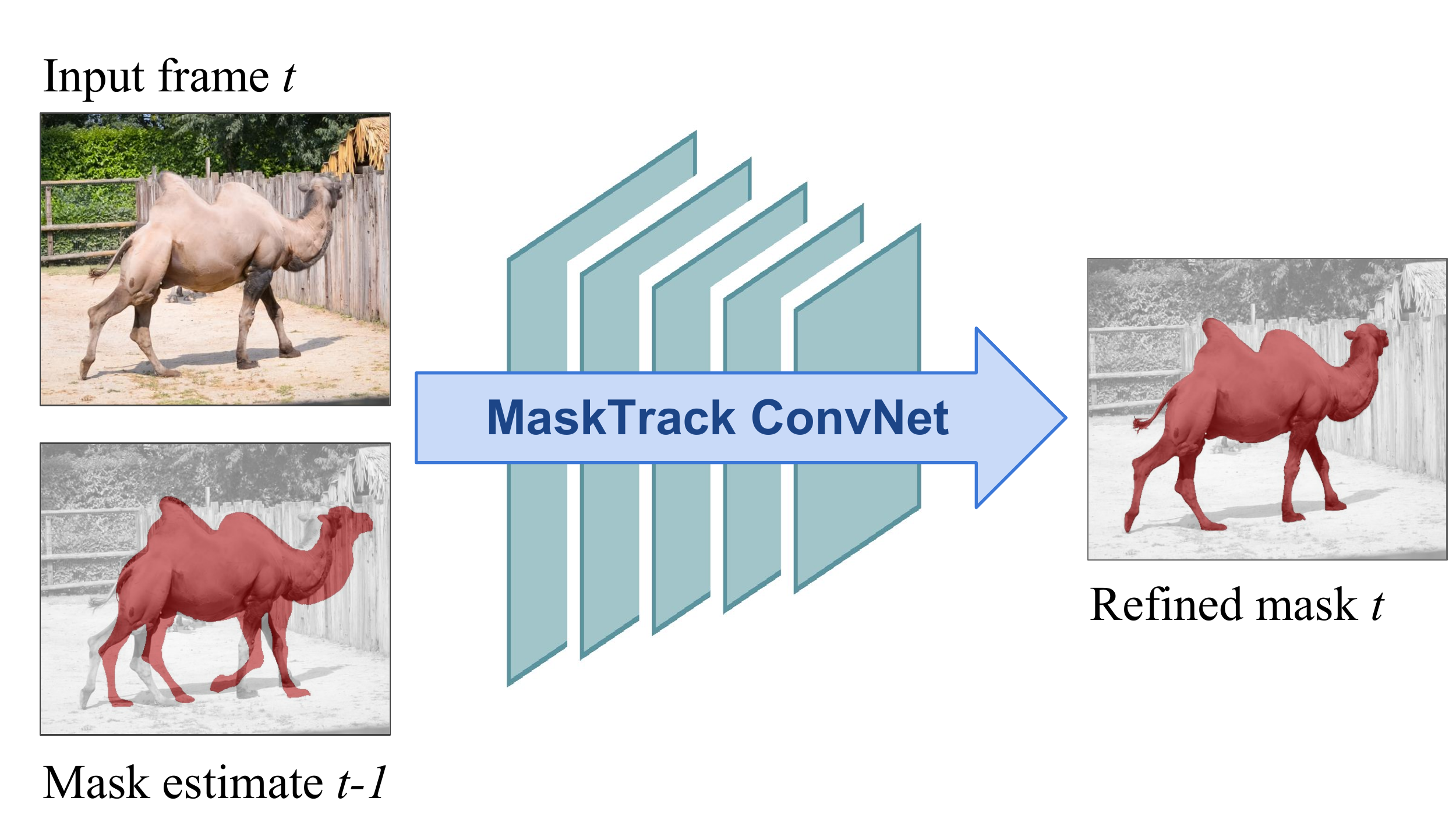}
\par\end{centering}
\caption{\label{fig:system-diagram}Given a rough mask estimate from the previous frame $t-1$
, we train a convnet to provide a refined mask output for the current frame $t$.}
\vspace{-1em}
\end{figure}

\begin{figure}
\begin{centering}
\hspace*{\fill}\subfloat[Annotated image]{\begin{centering}
\includegraphics[width=0.31\columnwidth,height=0.08\textheight]{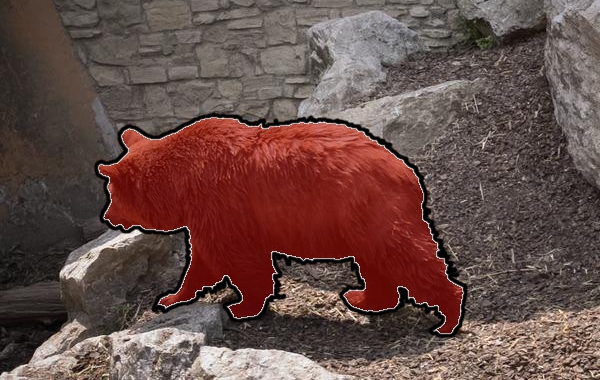}
\par\end{centering}
}\hspace*{\fill}\subfloat[Example training masks]{\begin{centering}
\includegraphics[width=0.31\columnwidth,height=0.08\textheight]{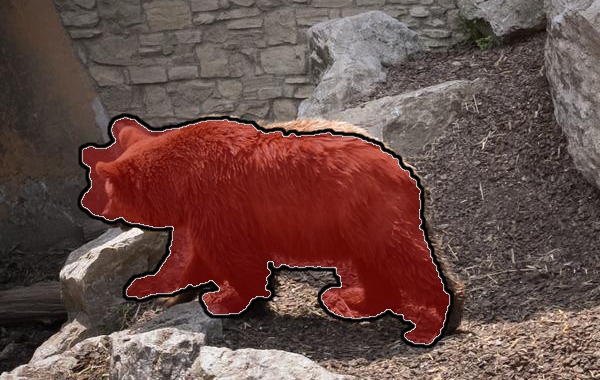}
\includegraphics[width=0.31\columnwidth,height=0.08\textheight]{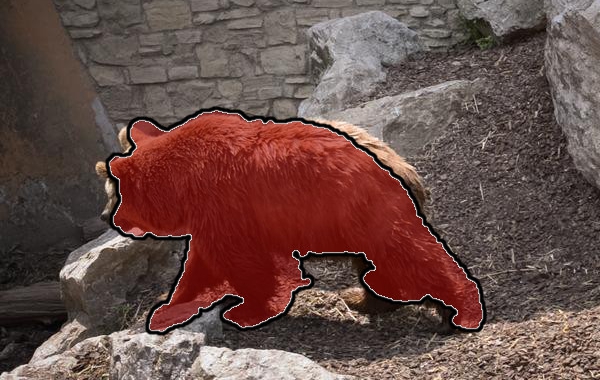}
\par\end{centering}
}\hspace*{\fill}
\par\end{centering}
\caption{\label{fig:offline-augmentation}Examples of training mask generation.
From one annotated image, multiple training masks are generated. The generated masks mimic plausible object shapes on the preceding frame.
}
\vspace{-1em}
\end{figure}

Figure \ref{fig:system-diagram} shows our overall architecture.
To simulate the noise in the previous frame output masks, during offline training we generate input masks by deforming the annotated masks via affine transformation as well as non-rigid deformations via thin-plate splines \cite{Bookstein1989Pami},
followed by a coarsening step (dilation morphological operation) to remove details of the object contour.
We apply this data generation procedure over a dataset of $\sim\negmedspace10^{4}$ images containing diverse object instances,
see examples in Figure \ref{fig:offline-augmentation}. At test time, given the mask estimate at time $t\negthinspace-\negthinspace1$, we apply the dilation operation and use the resulting rough mask as input for object
segmentation in frame $t$.

The affine transformations and non-rigid deformations aim at modelling the expected motion of an object between two frames. The coarsening permits us to generate training samples that resembles the test time data, simulating
the blobby shape of the output mask given from the previous frame by the convnet.
These two ingredients make the estimation more robust and help to avoid accumulation of errors from the preceding frames.

After training the resulting convnet has learnt to do guided instance segmentation, similar to networks like DeepMask \cite{Pinheiro2015Nips} and Hypercolumns \cite{Hariharan2015Cvpr}, but instead of taking a bounding box as guidance,
we can use an arbitrary input mask.
The training details are described in \S\ref{sec:convnet-details}.

When using offline training only, the segmentation procedure consists of two steps: the previous frame mask is coarsened and then fed into the trained network to estimate the current frame mask.
Since objects have a tendency to move smoothly through space, the object mask in the preceding frame will provide a good guess in the current frame
and simply copying the coarse mask from the previous frame is enough.
This approach is fast and already provides good results. We also experimented using optical flow to propagate the mask from one frame to the next, but found the optical flow errors to offset the gains.

With only the offline trained network, the proposed approach allows to achieve competitive performance compared to previously reported results (see \S\ref{sec:ablation-study}).
However the performance can be further improved by integrating online training strategy.

\subsection{Learning to segment instances online}
\label{sec:method-online}
For further boosting the video segmentation quality, we borrow and extend ideas that were originally proposed for tracking.
Current top performing tracking techniques \cite{Danelljan2016Eccv,Nam2016Cvpr} all use some form of online training. We thus consider improving results by adding this as a second strategy.

The idea is to use, at test time, the segment annotation of the first video frame as additional training data.
Using augmented versions of this single frame annotation, we proceed to fine-tune the model to become more specialized for the specific object instance at hand.

We use a similar data augmentation as for offline training. On top of affine and non-rigid deformations for the input mask, we also add image flipping and rotations to generate multiple training samples from one frame.
We generate $\sim\negmedspace10^3$ training samples from this single annotation,
and proceed to fine-tune the model previously trained offline.

With online fine-tuning, the network weights partially capture the appearance of the specific object being tracked. The model aims to strike a balance between general instance segmentation
(so as to generalize to the object changes), and specific instance segmentation (so as to leverage the common appearance across video frames). The details of the online fine-tuning are provided in
\S\ref{sec:convnet-details}. In our experiments we only do fine-tuning using the annotated frame(s).

To the best of our knowledge our approach is the first to use a pixel labelling network (like DeepLabv2 \cite{Chen2016ArxivDeeplabv2}) for the task of video object segmentation.
We name our full approach (using both offline and online training) $\mathtt{MaskTrack}$.

%

\subsection{Variants}
\label{sec:method-variants}
Additionally we consider variations of the proposed model.
First, we want to show that our approach is flexible and could handle different types of input annotations, using less supervision in the first frame annotation.
Second, motion information could be easily integrated in the system, improving the quality of the object segments.

\paragraph{Box annotation}
Here we discuss a variant named $\mathtt{MaskTrack}_{Box}$, that takes a bounding box annotation in the first frame as an input supervision instead of a segmentation mask.
To handle this variant we use on the first frame a second convnet model trained with bounding box rectangles as input masks. From the next frame onwards we use the standard $\mathtt{MaskTrack}$ model.

\begin{figure}
\begin{centering}
\begin{centering}
\begin{tabular}{@{}c@{ }c@{ }c@{ }c@{ }}
&\includegraphics[width=0.28\columnwidth,height=0.07\textheight]{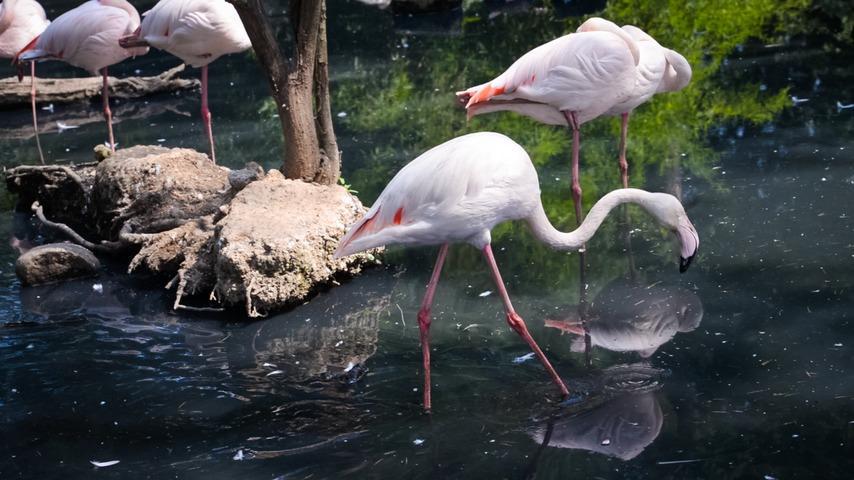} &
\includegraphics[width=0.28\columnwidth,height=0.07\textheight] {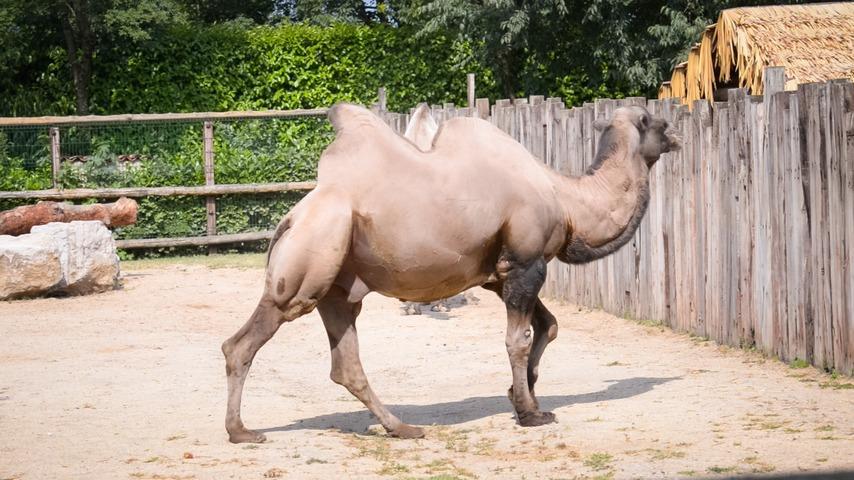} &
\includegraphics[width=0.28\columnwidth,height=0.07\textheight] {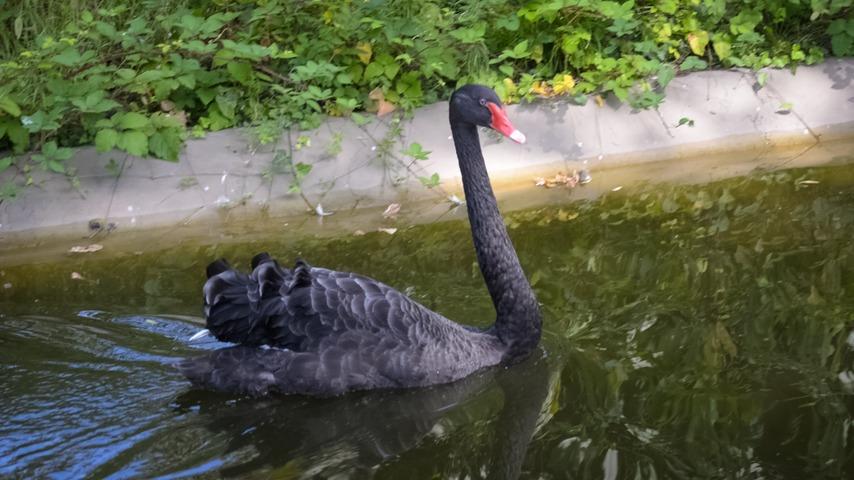} \tabularnewline
&\footnotesize{} & \footnotesize{} RGB Images & \tabularnewline
&\includegraphics[width=0.28\columnwidth,height=0.07\textheight]{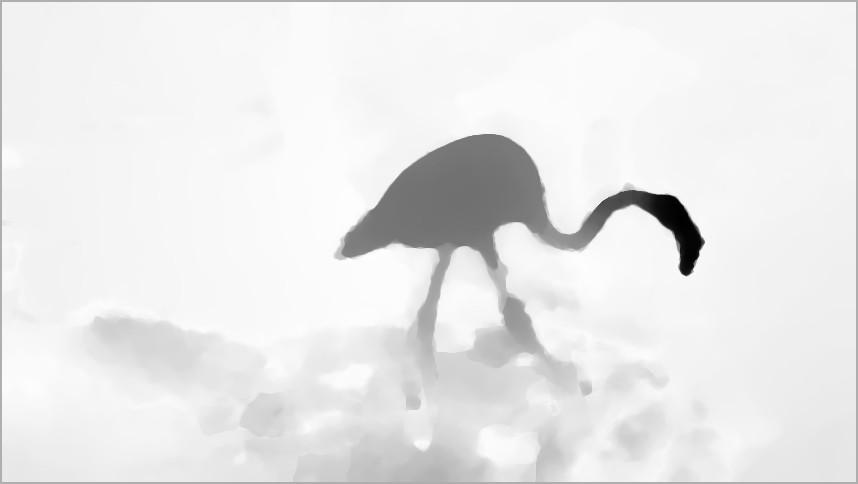} &
\includegraphics[width=0.28\columnwidth,height=0.07\textheight] {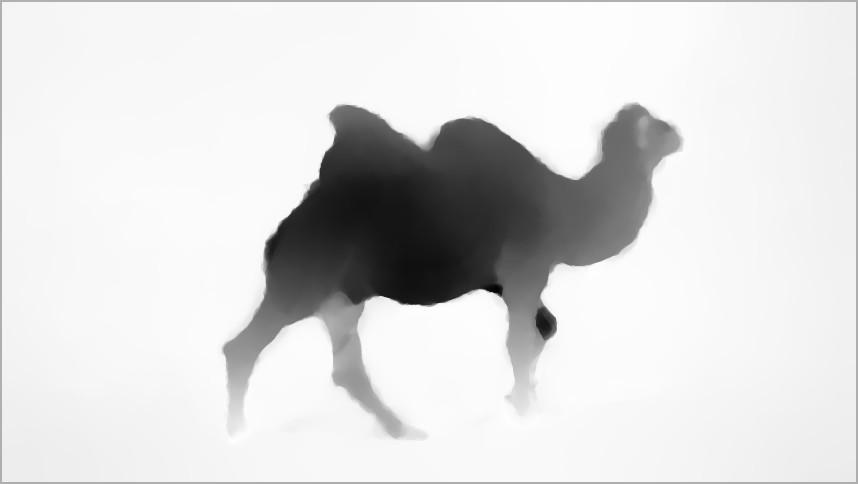} &
\includegraphics[width=0.28\columnwidth,height=0.07\textheight] {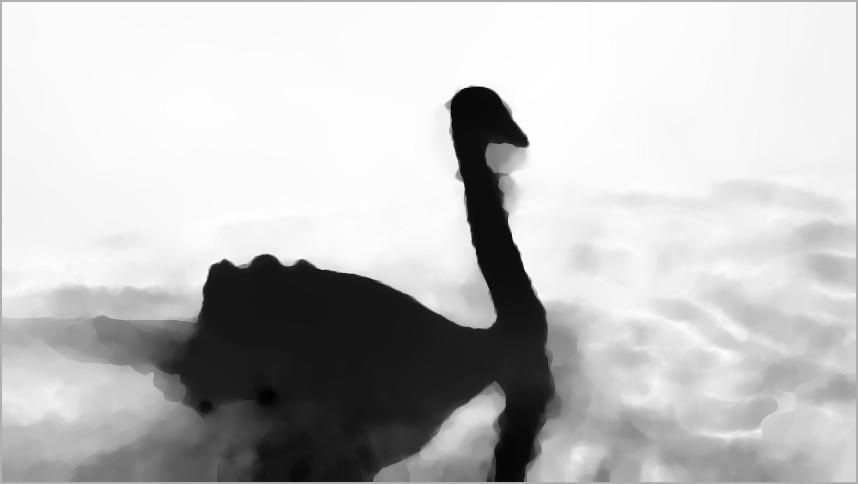}\tabularnewline
&\footnotesize{} & \footnotesize{} Optical flow magnitude & \tabularnewline
\end{tabular}
\par\end{centering}

\par\end{centering}
\caption{\label{fig:flow_images}Examples of optical flow magnitude images.}
\end{figure}

\paragraph{Optical flow}
On top of $\mathtt{MaskTrack}$, we consider to employ optical flow as a source of additional information to guide the segmentation.
Given a video sequence, we compute the optical flow using EpicFlow \cite{EpicFlowCVPR15} with Flow Fields matches \cite{FlowFields15} and convolutional boundaries \cite{COB_Maninis16}.
In parallel to the vanilla $\mathtt{MaskTrack}$, we proceed to compute a second output mask using the magnitude of the optical flow field as input image (replicated into a three channel image).
The model is used as-is, without retraining.
Although it has been trained on RGB images, this strategy works because object flow magnitude roughly looks like a gray-scale object, and still captures useful object shape information, see examples in Figure \ref{fig:flow_images}.
Using the RGB model allows to avoid training the convnet on a video dataset with segmentation annotations.

We then fuse by averaging the output scores given by the two parallel networks (using RGB image and optical flow magnitude as inputs). We name this variant $\mathtt{MaskTrack\mathsmaller{+}Flow}$.
Optical flow provides complementary information to the $\mathtt{MaskTrack}$ with RGB images, improving the overall performance.

\section{Network implementation and training}
\label{sec:convnet-details}
Following, we describe the implementation details of our approach, namely the offline and online training strategy and the data augmentation.
\paragraph{Network}
For all our experiments we use the training and test parameters of DeepLabv2-VGG network \cite{Chen2016ArxivDeeplabv2}. The model is initialized from a VGG16 network pre-trained on ImageNet \cite{Simonyan2015Iclr}.
For the extra mask channel of filters in the first convolutional layer we use gaussian initialization. We also tried zero initialization, but observed no difference.

\paragraph{Offline training}
The advantage of our method is that it does not require expensive pixel-label annotations on videos for training the convnet. Thus we can employ images and annotations from existing saliency segmentation datasets.
We consider images and segmentation masks from ECSSD \cite{ShiYXJ16}, MSRA10K \cite{ChengPAMI}, SOD \cite{Movahedi2010DesignAP}, and PASCAL-S \cite{LiHKRY14}.
This results in an aggregate set of $11\,282$ training images.

The input masks for an extra channel are generated by deforming the binary segmentation masks via affine transformation and non-rigid deformations, as discussed in \S\ref{sec:method-offline}.
For affine transformation we consider random scaling ($\pm5\%$ of object size) and translation ($\pm10\%$ shift).
Non-rigid deformations are done via thin-plate splines \cite{Bookstein1989Pami} using $5$ control points and randomly shifting the points in x and y directions within $\pm10\%$ margin of the original segmentation mask width and height.
Then the mask is coarsened using dilation operation with $5$ pixel radius.
This mask deformation procedure is applied over all object instances in the training set. For each image two different masks are generated, see examples in Figure \ref{fig:offline-augmentation}.

For training we follow \cite{Chen2016ArxivDeeplabv2} and use SGD with mini-batches of $10$ images and a polynomial learning policy with initial learning rate of $0.001$. The momentum and weight decay are set to $0.9$ and $0.0005$,
respectively.
The network is trained for $20\text{k}$ iterations.

\paragraph{Online training}
For online adaptation we finetune the model previously trained offline on the first frame for 200 iterations with training samples generated from the first frame annotation.
We augment the first frame by image flipping and rotations as well as by deforming the annotated masks for an extra channel via affine and non-rigid deformations with the same parameters as for the offline training.
This results in an augmented set of $\sim\negmedspace10^3$ training images.

The network is trained with the same learning parameters as for offline training, finetuning all convolutional and fully connected layers.

At test time our base $\mathtt{MaskTrack}$ system runs at about $12$ seconds per frame (average over DAVIS dataset, amortizing the online fine-tuning time over all video frames), which is a magnitude faster compared to
$\mathtt{ObjFlow}$ \cite{Tsai2016Cvpr} (takes $2$ minutes per frame, averaged over DAVIS dataset).


\section{Results}
\label{sec:results}

In this section we
describe our evaluation protocol (\S\ref{sec:experimental-setup}),
study the quantitative importance of the different components of our system (\S\ref{sec:ablation-study}), and report results comparing to state of art techniques over three datasets
(190 videos total, \S\ref{sec:single-frame-results}), as well as comparing the effects of different amounts of annotation on final quality (\S\ref{sec:few-frames-results}).
Additional quantitative and qualitative results are provided in the supplementary material.

\subsection{Experimental setup}
\label{sec:experimental-setup}


\paragraph{Datasets}
We evaluate the proposed approach on three different video object segmentation datasets: DAVIS \cite{Perazzi2016Cvpr}, YoutubeObjects \cite{Prest2012Cvpr}, and \segtrack{} \cite{Li2013Iccv}.
These datasets include assorted challenges such as appearance change, occlusion, motion blur and shape deformation.

DAVIS \cite{Perazzi2016Cvpr} consists of $50$ high quality videos, totalling $3\,455$ frames. Pixel-level segmentation annotations are provided for each frame, where one single object or two connected objects are
separated from the background.

YoutubeObjects \cite{Prest2012Cvpr} includes videos with 10 object categories. We consider the subset of $126$ videos with more than $20\,000$ frames, for which the pixel-level ground truth segmentation masks are provided
by \cite{Jain2014Eccv}.

\segtrack{} \cite{Li2013Iccv} contains $14$ video sequences with $24$ objects and $947$ frames. Every frame is annotated with a pixel-level object mask.
As instance-level annotations are provided for sequences with multiple objects, each specific instance segmentation is treated as separate problem.

\paragraph{Evaluation}
We evaluate using the standard mIoU metric: intersection-over-union of the estimated segmentation and the ground truth binary mask, also known as \emph{Jaccard Index}, averaged across videos.
For DAVIS we use the provided benchmark code \cite{Perazzi2016Cvpr}, which excludes the first and the last frames from the evaluation. For YoutubeObjects and \segtrack{} only the first frame is excluded.

Previous work used diverse evaluations procedure. To ensure a consistent comparison between methods, when needed, we re-computed scores for using shared results maps, or by generating them using available open source code.
In particular, we collected new results for $\mathtt{ObjFlow}$ \cite{Tsai2016Cvpr} and $\mathtt{BVS}$ \cite{Maerki2016Cvpr} in order to present other methods with results across the three datasets.

\begin{figure}
\begin{centering}
\begin{centering}
\begin{tabular}{@{}c@{ }c@{ }c@{ }c@{ }}
\begin{turn}{90}
{\footnotesize{} \hspace{0.5em} 1st frame}
\end{turn}
&\includegraphics[width=0.3\columnwidth,height=0.08\textheight]{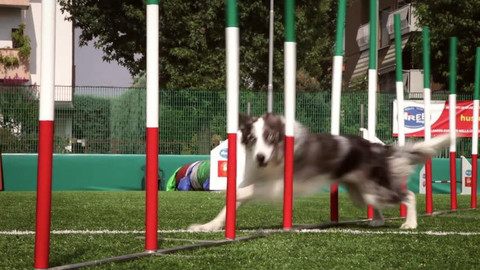} &
\includegraphics[width=0.3\columnwidth,height=0.08\textheight]{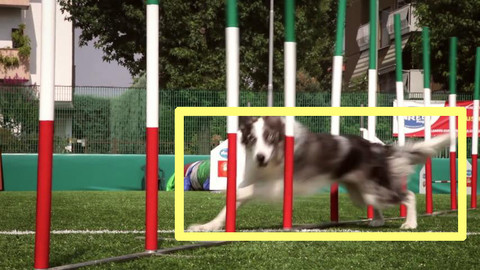} &
\includegraphics[width=0.3\columnwidth,height=0.08\textheight]{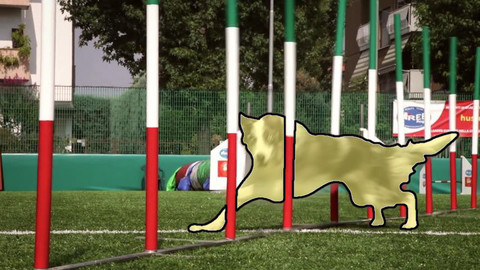} \tabularnewline
&\footnotesize{}Image & \footnotesize{}Box annotation & \footnotesize{}Segment annotation\tabularnewline
\begin{turn}{90}
{\footnotesize{} \hspace{0.5em} 13th frame}
\end{turn}
&\includegraphics[width=0.3\columnwidth,height=0.08\textheight]{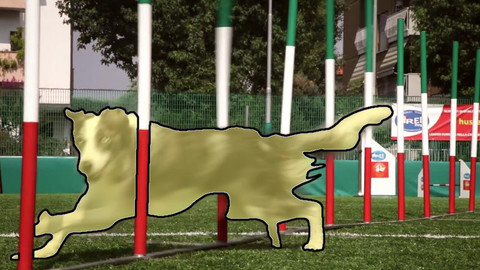} &
\includegraphics[width=0.3\columnwidth,height=0.08\textheight]{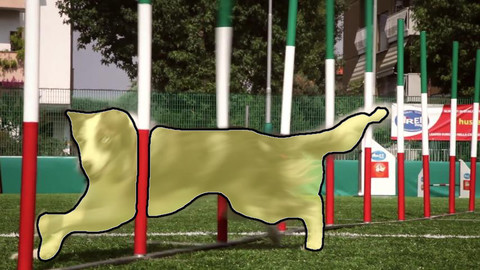} &
\includegraphics[width=0.3\columnwidth,height=0.08\textheight]{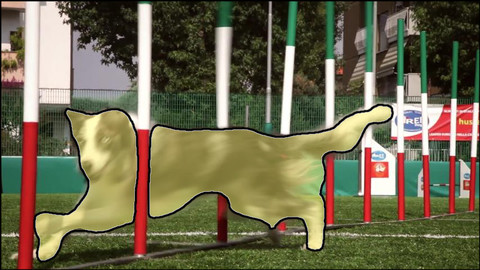}\tabularnewline
&\footnotesize{}Ground truth & \footnotesize{} $\mathtt{MaskTrack}_{Box}$ result  & \footnotesize{} $\mathtt{MaskTrack}$ result\tabularnewline
\end{tabular}
\par\end{centering}

\par\end{centering}
\caption{\label{fig:teaser}By propagating annotation from the 1st frame, either from segment or just bounding box annotations, our system generates results comparable to ground truth.}
\end{figure}

\subsection{Ablation study}
\label{sec:ablation-study}

We first study different ingredients of our method. We experiment on the DAVIS dataset and measure the performance using the mean intersection-over-union metric (mIoU).
Table \ref{tab:ablation-study} shows the importance of each of the ingredients described in \S\ref{sec:method} and reports the improvement of adding extra components to the $\mathtt{MaskTrack}$ model.

\paragraph{Add-ons}
We first study the effect of adding a couple of ingredients on top of our base $\mathtt{MaskTrack}$ system, which are specifically fine-tuned for DAVIS.
We see that optical flow provides complementary information to the appearance, boosting further the results ($74.8 \rightarrow 78.4$). Adding on top a well-tuned post-processing CRF \cite{Kraehenbuehl2011Nips}
can gain a couple of mIoU points, reaching $80\%\ \text{mIoU}$ on DAVIS, the best known result on this dataset.

Albeit optical flow can provide interesting gains, we found it to be brittle when going across different datasets. Different strategies to handle optical flow provide $1\negmedspace\sim\negmedspace4 \%$
on each dataset,
but none provide consistent gains across all datasets; mainly due to failure modes of the optical flow algorithms. For the sake of presenting a single model with fix parameters across all datasets, we refrain from using
a per-dataset tuned optical flow in the results of \S\ref{sec:single-frame-results}.

\paragraph{Training}
We next study the effect of offline/online training of the network.
By disabling online fine-tuning, and only relying on offline training we see a $\sim\negmedspace5$ IoU percent points drop, showing that online fine-tuning indeed expand the tracking capabilities.
If instead we skip offline training and only rely on online fine-tuning performance drop drastically, albeit the absolute quality ($57.6\ \text{mIoU}$) is surprisingly high for a system trained on ImageNet+single frame.

By reducing the amount of training data from $11\text{k}$ to $5\text{k}$ we only see a minor decrease in mIoU; this indicates that the amount of training data is not critical to get reasonable performance.
That being said, further increase of the training data volume would lead to improved results.

Additionally, we explore the effect of the offline training on video data instead of using static images. We train the model on the annotated frames of two combined datasets, \segtrack{} and YoutubeObjects.
By switching to train on video data we observe a minor decrease in mIoU; this could be explained by lack of diversity in the video training data due to the small scale of the existing datasets,
as well as the effect of the domain shift between different benchmarks. This shows that employing static images in our approach does not result in any performance drop.
\paragraph{Mask deformation}
We also study the influence of mask deformations.
We see that coarsening the mask via dilation provides a small gain
, as well as adding non-rigid deformations.
All-and-all, Table \ref{tab:ablation-study} shows that the main factor affecting the quality is using any form of mask deformations when creating the training samples (both for offline and online training).
This ingredient is critical for our overall approach, making the segmentation estimation more robust at test time to the noise in the input mask.

\paragraph{Input channel}
Next we experiment with different variants  of the extra channel input.
Even by changing the input from segments to boxes, a model trained for this modality still provides reasonable results.

Most interestingly we also evaluated a model that does not use any mask input. Without the additional input channel, this pixel labelling convnet was trained for saliency offline and fine-tuned online to capture the appearance
of the object of interest. This model obtains competitive results ($72.5\ \text{mIoU}$), showing the power of modelling the video segmentation task as an instance segmentation problem.


\begin{table}
\begin{centering}
\begin{tabular}{@{  }c@{  }l@{  }c@{  }c@{  }}
Aspect & System variant  & mIoU  & $\Delta\text{mIoU}$\tabularnewline
\hline
\hline
\multirow{2}{*}{Add-ons}
& $\mathtt{MaskTrack\mathsmaller{+}Flow \mathsmaller{+}CRF}$  & \textbf{80.3} & $+1.9$\tabularnewline
& $\mathtt{MaskTrack\mathsmaller{+}Flow}$  & $78.4$ & $+3.6$\tabularnewline
 \hline
& $\mathtt{MaskTrack}$ & \textit{74.8} & \hspace{1.5em}-\tabularnewline
\hline
\multirow{4}{*}{\begin{tabular}{c}Training \end{tabular}}
 & No online fine-tuning  & $69.9$ & $-4.9$\tabularnewline
 & No offline training  & $57.6$ & $-17.2$\tabularnewline
 & Reduced offline training  & $73.2$ & $-1.6$\tabularnewline
 & Training on video& $72.0$ & $-2.8$\tabularnewline
\arrayrulecolor{lightgray}\hline
\multirow{3}{*}{%
\begin{tabular}{c}
Mask\tabularnewline
defor-\tabularnewline
mation\tabularnewline
\end{tabular}}
 & No dilation  & $72.4$ & $-2.4$\tabularnewline
 & No deformation  & $17.1$ & $-57.7$\tabularnewline
 & No non-rigid deformation & $73.3$ & $-1.5$\tabularnewline
\arrayrulecolor{lightgray}\hline
\multirow{2}{*}{%
\begin{tabular}{c}
Input\tabularnewline
channel\tabularnewline
\end{tabular}}
 & Boxes  & $69.6$ & $-5.2$\tabularnewline
 & No input  & $72.5$ & $-2.3$\tabularnewline

\end{tabular}
\par\end{centering}
\caption{\label{tab:ablation-study}Ablation study of our $\mathtt{MaskTrack}$
method on DAVIS. Given our full system, we change one
ingredient at a time, to see each individual contribution. See \S\ref{sec:ablation-study} for discussion.}
\end{table}

\begin{table}
\begin{centering}
\begingroup
\begin{tabular}{l|ccc}
\multirow{2}{*}{Method} & \multicolumn{3}{c}{Dataset, mIoU}\tabularnewline
	& {\footnotesize{}DAVIS} & {\footnotesize{}YoutbObjs} & {\footnotesize{}\segtrack{}}\tabularnewline
\hline
\hline
Box oracle  & 45.1  & 55.3 & 56.1 \tabularnewline
Grabcut oracle & 67.3 & 67.6 & 74.2\tabularnewline
\hline
$\mathtt{ObjFlow}$ \cite{Tsai2016Cvpr} & 71.4 & 70.1 & 67.5\tabularnewline
$\mathtt{BVS}$ \cite{Maerki2016Cvpr} & 66.5 & 59.7 & 58.4\tabularnewline
 $\mathtt{NLC}$ \cite{Faktor2014Bmvc} & 64.1 & - & -\tabularnewline
$\mathtt{FCP}$ \cite{Perazzi2015Iccv} & 63.1 & - & -\tabularnewline
$\mathtt{W16}$ \cite{Wang2016Accv} & - & 59.2 & -\tabularnewline
$\mathtt{Z15}$ \cite{Zhang2015CvprObjectSegmentation} & - & 52.6 & -\tabularnewline
$\mathtt{TRS}$ \cite{Xiao2016Cvpr}  & -  & - & \textbf{69.1} \tabularnewline
\hline
$\mathtt{MaskTrack}$\hspace*{1.5em} & \textbf{74.8} & \textbf{71.7} & 67.4\tabularnewline
$\mathtt{MaskTrack}_{Box}$ & 73.7  & 69.3 & 62.4\tabularnewline
\end{tabular}\endgroup
\par\end{centering}
\caption{\label{tab:VOS-results}Video object segmentation results on three
datasets. Compared to related state of the art, our approach provides
consistently good results. On DAVIS the extended version of our system $\mathtt{MaskTrack\mathsmaller{+}Flow \mathsmaller{+}CRF}$ reaches $80.3\ \text{mIoU}$. See \S\ref{sec:single-frame-results} for details.}
\end{table}

\subsection{Single frame annotations}
\label{sec:single-frame-results}

Table \ref{tab:VOS-results} presents results when the first frame is annotated with an object segmentation mask. This is the protocol commonly used on DAVIS, \segtrack{}, and YoutubeObjects.

We see that $\mathtt{MaskTrack}$ obtains competitive performance across all three datasets. This is achieved using our purely frame-by-frame feed-forward system, using the exact same model and parameters across all datasets.
Our $\mathtt{MaskTrack}$ results are obtained in a single pass, do not use any global optimization, not even optical flow.
We believe this shows the promise of formulating video object segmentation from the instance segmentation perspective.

On \segtrack{}, $\mathtt{JOTS}$ \cite{Wen2015Cvpr} reported higher numbers ($71.3\ \text{mIoU}$), however, they report tuning their method parameters' per video, and thus it is not comparable to our setup with fix-parameters.

Table \ref{tab:VOS-results} also reports results for the $\mathtt{MaskTrack}_{Box}$ variant described in \S\ref{sec:method-variants}.
Starting only from box annotations on the first frame, our system still generates comparably good results (see Figure \ref{fig:teaser}), remaining on the top three best results in all the datasets covered.

By adding additional ingredients specifically tuned for different datasets, such as optical flow (see \S\ref{sec:method-variants}) and CRF post-processing, we can push the results even further, reaching
$80.3\ \text{mIoU}$ on DAVIS, $72.6$ on YoutubeObjects and  $70.3$ on \segtrack{}. The dataset specific tuning is described in the supplementary material.

Figure \ref{fig:qualitative-results} presents qualitative results of the proposed $\mathtt{MaskTrack}$ model across three different datasets.

\begin{figure}
\begin{centering}
\includegraphics[width=1.0\columnwidth]{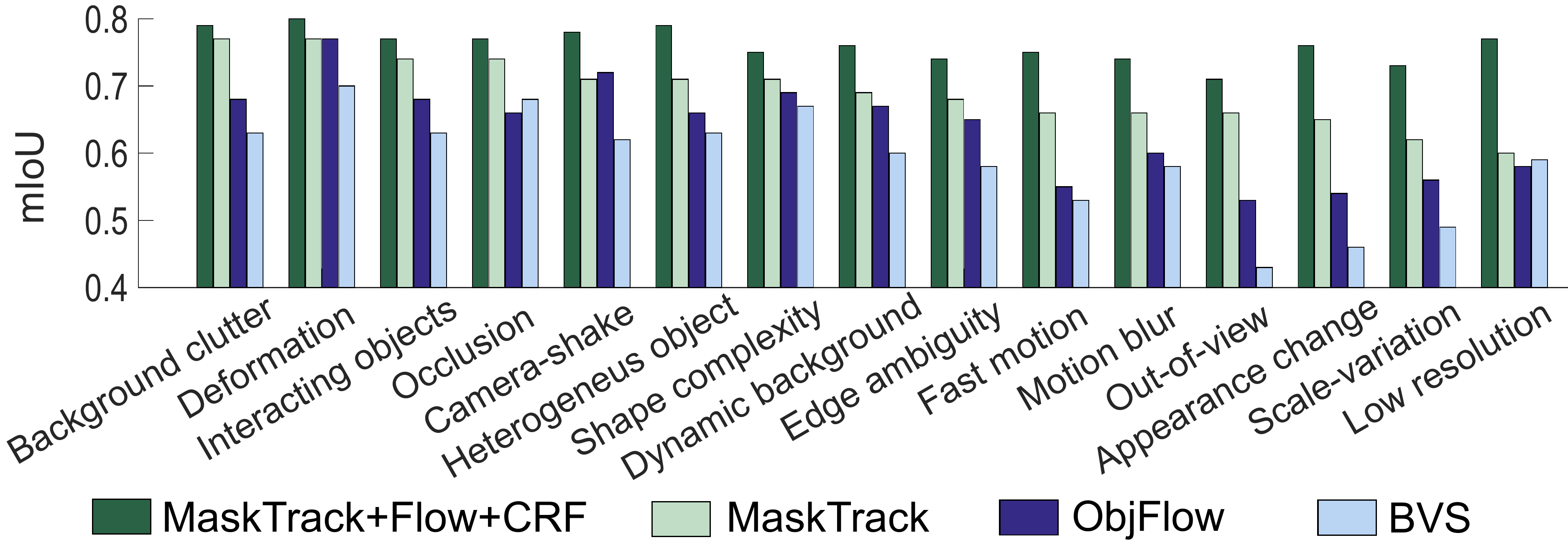}
\par\end{centering}
\caption{\label{fig:attribute_analysis} Attribute based evaluation on DAVIS.}
\end{figure}

\paragraph{Attribute-based analysis}
Figure \ref{fig:attribute_analysis} presents a more detailed evaluation on DAVIS \cite{Perazzi2016Cvpr} using video attributes.
The attribute based analysis shows that our generic model, $\mathtt{MaskTrack}$,
is robust to various video challenges present in DAVIS. It compares favourably on any subset of videos sharing the same attribute, except camera-shake, where $\mathtt{ObjFlow}$ \cite{Tsai2016Cvpr} marginally outperforms our approach.
We observe that $\mathtt{MaskTrack}$ handles fast-motion and motion-blur well, which are
typical failure cases for methods relying on spatio-temporal connections
\cite{Maerki2016Cvpr, Tsai2016Cvpr}.

Due to the online fine-tuning on the first frame annotation of a new video, our system is able to capture the appearance of the specific object of interest. 
This allows it to better recover from occlusions, out-of-view scenarios and appearance changes, which usually affect methods that strongly rely on propagating segmentations on a per-frame basis.

Incorporating optical flow information into $\mathtt{MaskTrack}$ substantially increases
robustness on all categories. As one could expect, $\mathtt{MaskTrack\mathsmaller{+}Flow \mathsmaller{+}CRF}$ better discriminates cases
involving color ambiguity and salient motion. However, we also observed less-obvious improvements in cases with scale-variation and low-resolution objects.

\paragraph{Conclusion}
With our simple, generic system for video object segmentation we are able to achieve competitive results with existing techniques, on three different datasets.
These results are obtained with fixed parameters, from a forward-pass only, using only static images during offline training. We also reach good results even when using only a box annotation as starting point.

\begin{figure*}
\begin{centering}
\setlength{\tabcolsep}{1.5pt}
\begin{tabular}{cccccc}

\multirow{2}{*}{\begin{turn}{90}
{\footnotesize{}DAVIS}
\end{turn}} &
\includegraphics[width=0.16\textwidth,height=0.06\textheight]{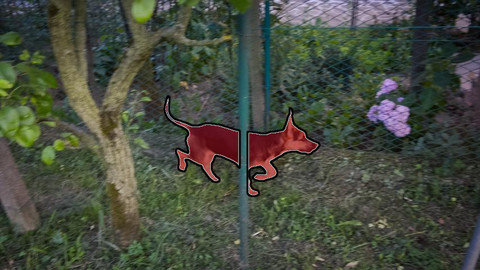} &
\includegraphics[width=0.16\textwidth,height=0.06\textheight]{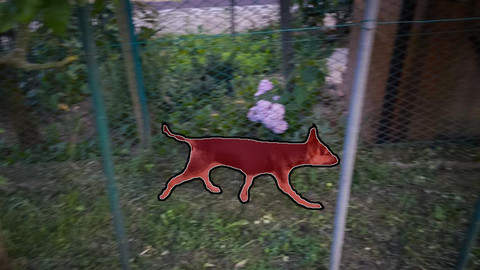} &
\includegraphics[width=0.16\textwidth,height=0.06\textheight]{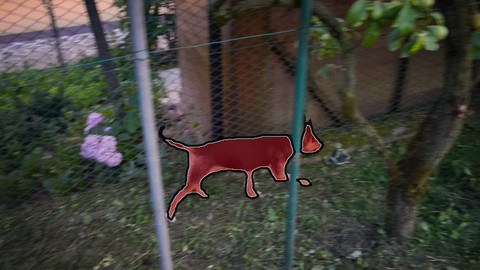} &
\includegraphics[width=0.16\textwidth,height=0.06\textheight]{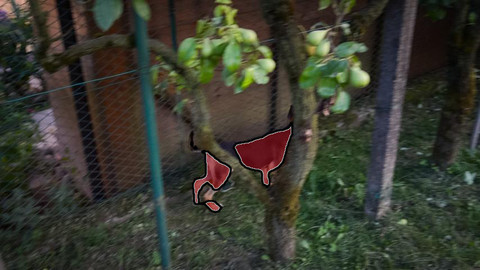} &
\includegraphics[width=0.16\textwidth,height=0.06\textheight]{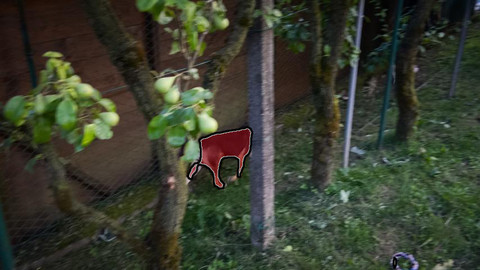} \tabularnewline
 &
\includegraphics[width=0.16\textwidth,height=0.06\textheight]{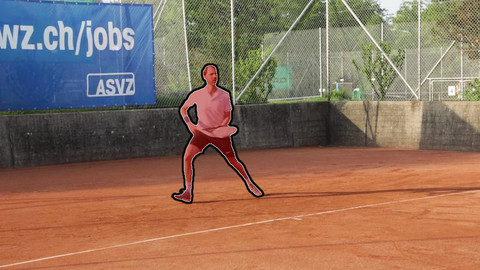} &
\includegraphics[width=0.16\textwidth,height=0.06\textheight]{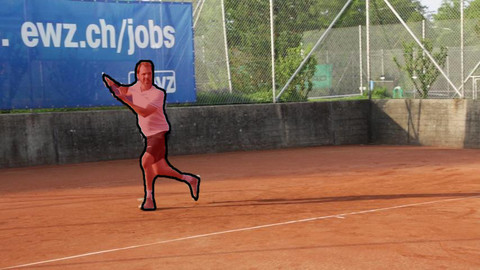} &
\includegraphics[width=0.16\textwidth,height=0.06\textheight]{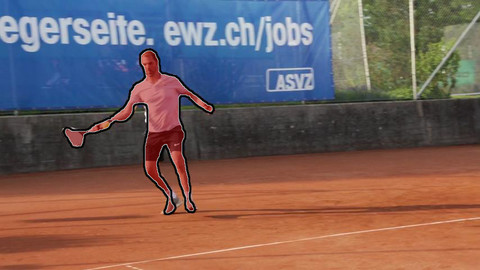} &
\includegraphics[width=0.16\textwidth,height=0.06\textheight]{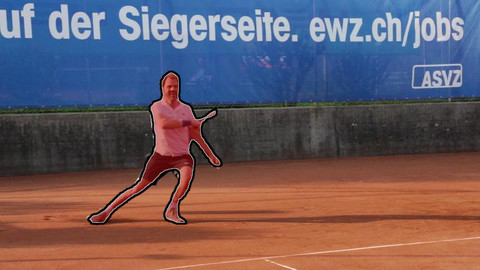} &
\includegraphics[width=0.16\textwidth,height=0.06\textheight]{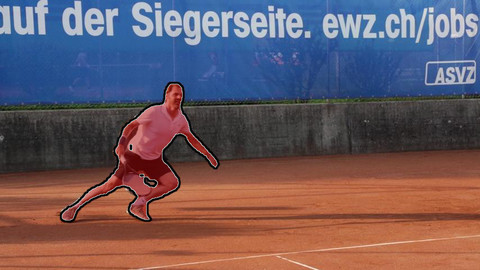} \tabularnewline
 \multirow{2}{*}{\begin{turn}{90}
{\footnotesize{}SegTrack}
\end{turn}}
 &
\includegraphics[width=0.16\textwidth,height=0.06\textheight]{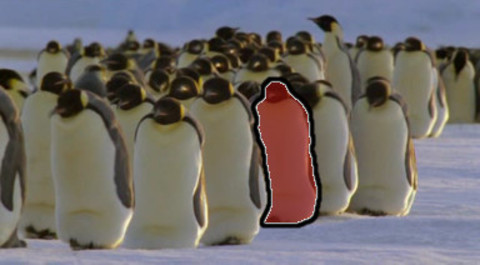} &
\includegraphics[width=0.16\textwidth,height=0.06\textheight]{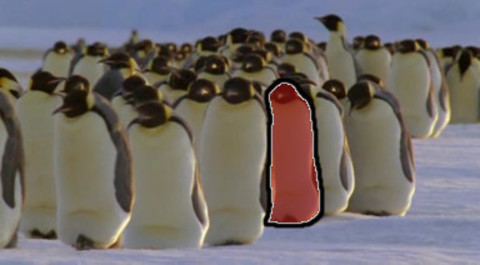} &
\includegraphics[width=0.16\textwidth,height=0.06\textheight]{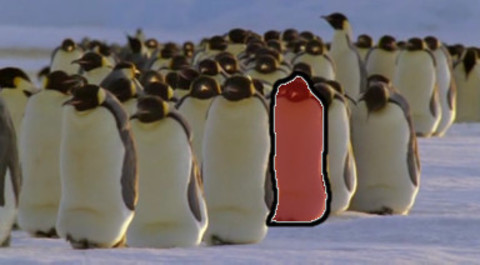} &
\includegraphics[width=0.16\textwidth,height=0.06\textheight]{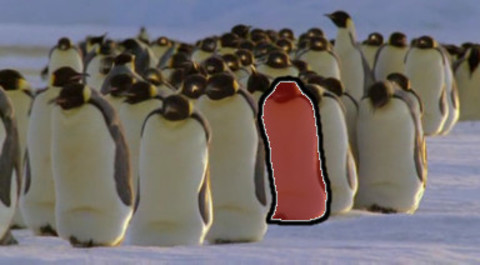} &
\includegraphics[width=0.16\textwidth,height=0.06\textheight]{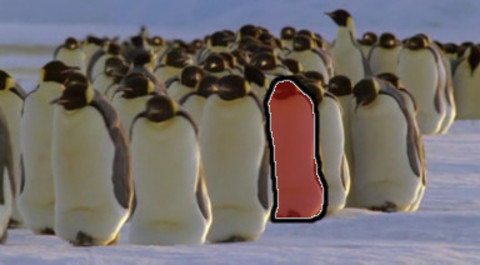} \tabularnewline
&
\includegraphics[width=0.16\textwidth,height=0.06\textheight]{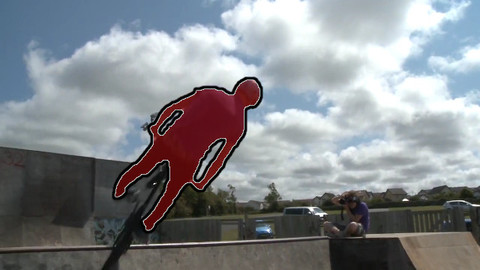} &
\includegraphics[width=0.16\textwidth,height=0.06\textheight]{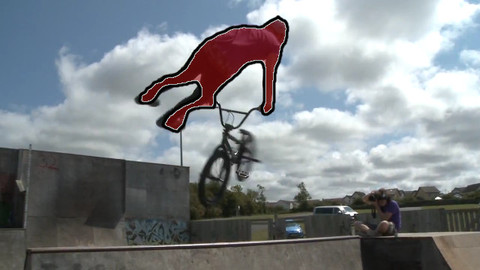} &
\includegraphics[width=0.16\textwidth,height=0.06\textheight]{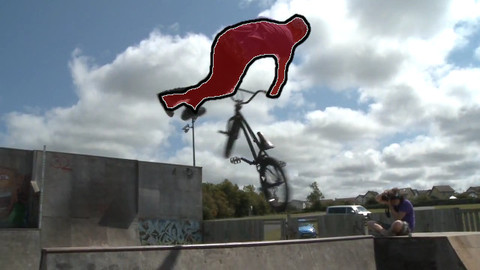} &
\includegraphics[width=0.16\textwidth,height=0.06\textheight]{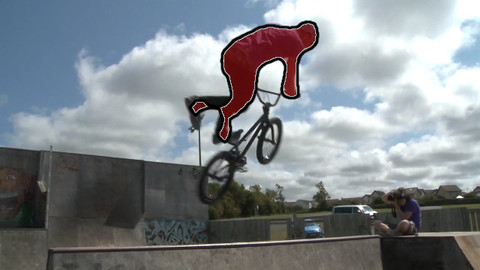} &
\includegraphics[width=0.16\textwidth,height=0.06\textheight]{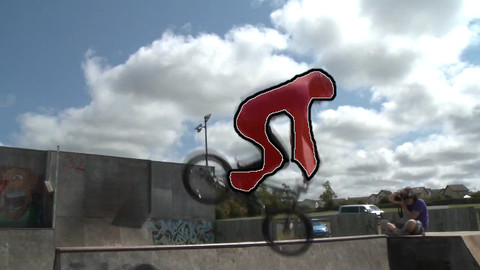} \tabularnewline

 \multirow{2}{*}{\begin{turn}{90}
{\footnotesize{}YouTube}
\end{turn}} &
\includegraphics[width=0.16\textwidth,height=0.06\textheight]{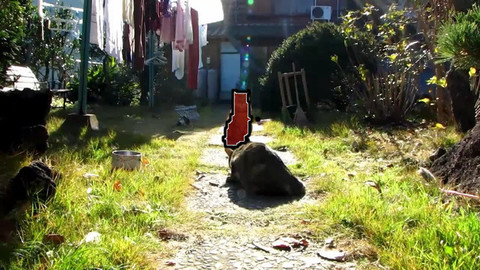} &
\includegraphics[width=0.16\textwidth,height=0.06\textheight]{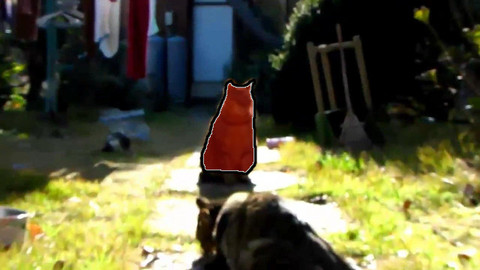} &
\includegraphics[width=0.16\textwidth,height=0.06\textheight]{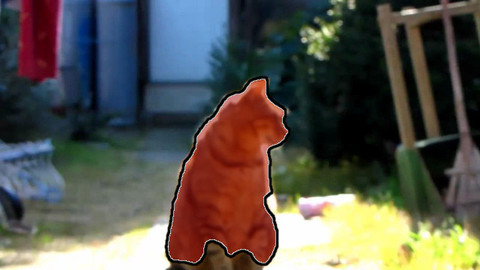} &
\includegraphics[width=0.16\textwidth,height=0.06\textheight]{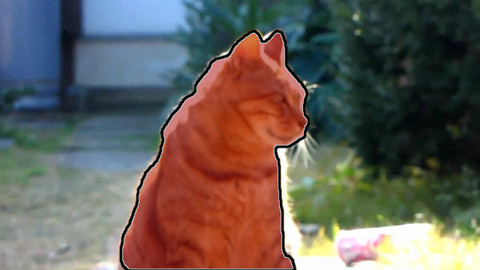} &
\includegraphics[width=0.16\textwidth,height=0.06\textheight]{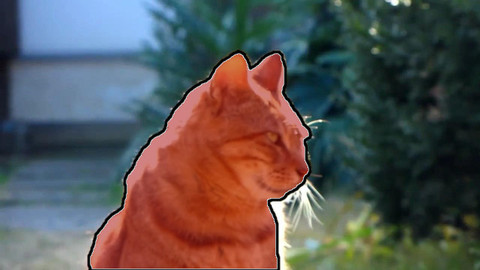} \tabularnewline
   &
\includegraphics[width=0.16\textwidth,height=0.06\textheight]{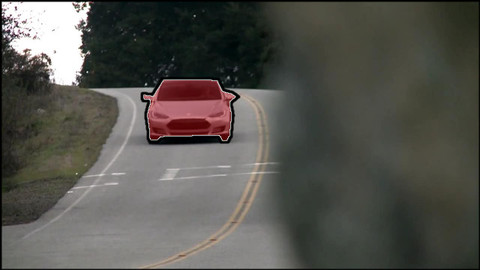} &
\includegraphics[width=0.16\textwidth,height=0.06\textheight]{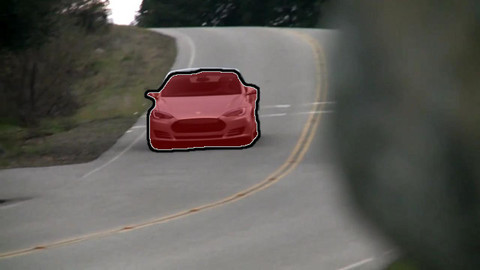} &
\includegraphics[width=0.16\textwidth,height=0.06\textheight]{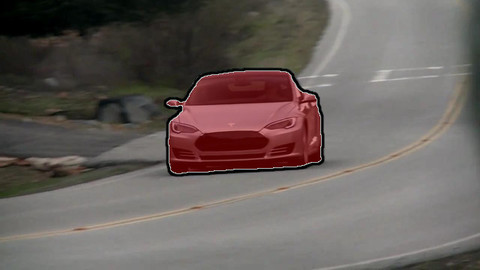} &
\includegraphics[width=0.16\textwidth,height=0.06\textheight]{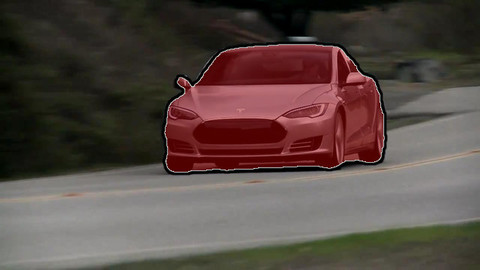} &
\includegraphics[width=0.16\textwidth,height=0.06\textheight]{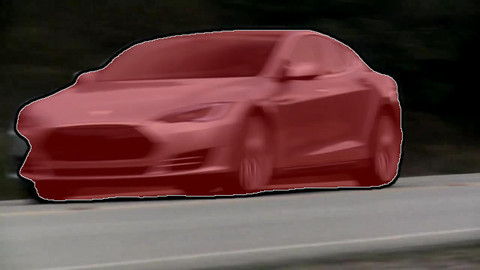} \tabularnewline
&\footnotesize{}1st frame, GT segment & \multicolumn{4}{c}{\footnotesize{} Results with $\mathtt{MaskTrack}$, the frames are chosen equally distant based on the video sequence length}
\end{tabular}
\par\end{centering}
\caption{\label{fig:qualitative-results}
Qualitative results of three different datasets. Our algorithm is robust to challenging situations such as occlussions, fast motion, multiple instances of the same semantic class,
object shape deformation, camera view change and motion blur.}
\end{figure*}

\subsection{Multiple frames annotations}
\label{sec:few-frames-results}

\begin{figure}
\hspace*{\fill}\subfloat[\label{fig:segment-annotations-density}Segment annotations]{\begin{centering}
\includegraphics[width=0.8\columnwidth]{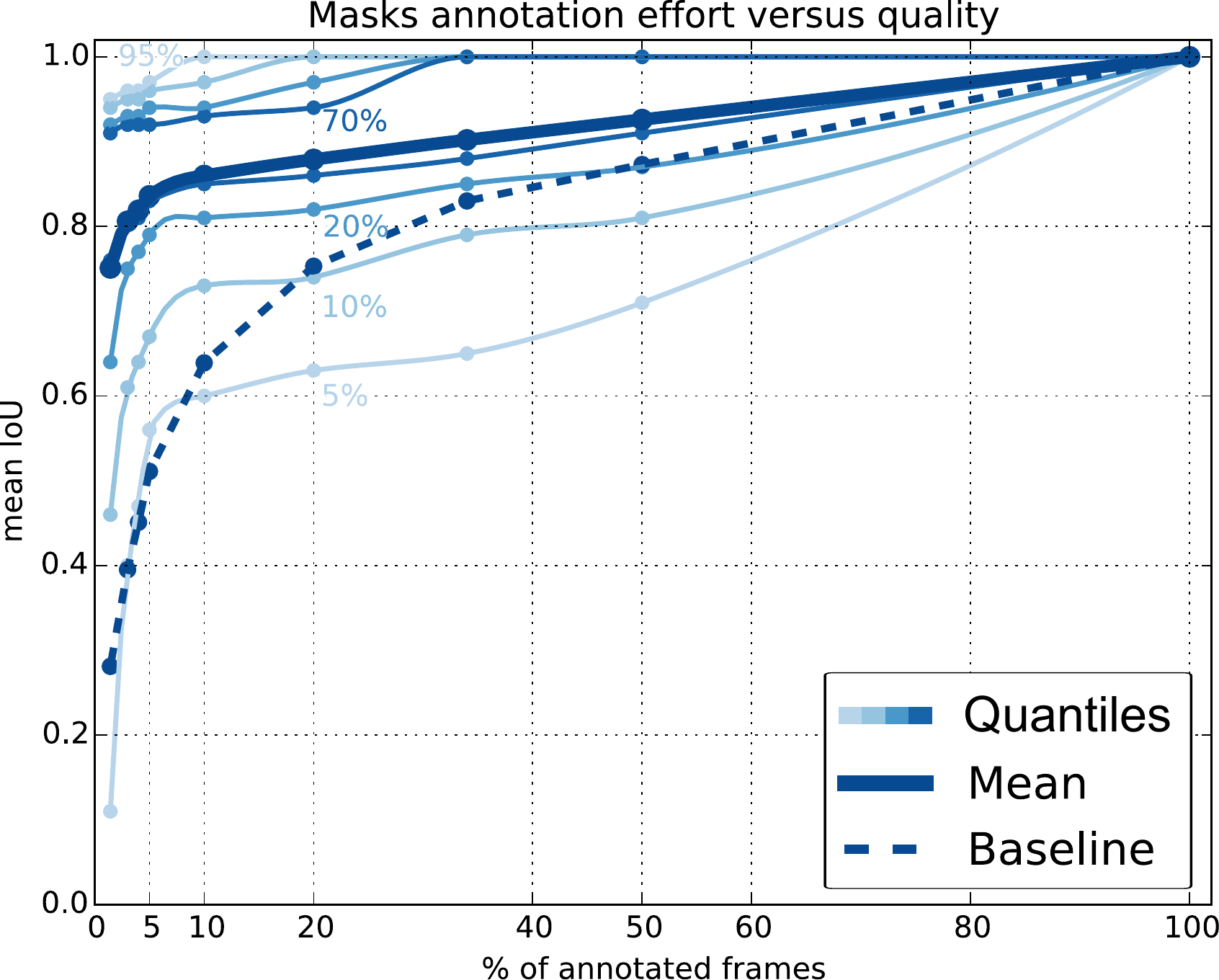}
\par\end{centering}
}\hspace*{\fill}

\hspace*{\fill}\subfloat[\label{fig:box-annotations-density}Box annotations]{\begin{centering}
\includegraphics[width=0.8\columnwidth]{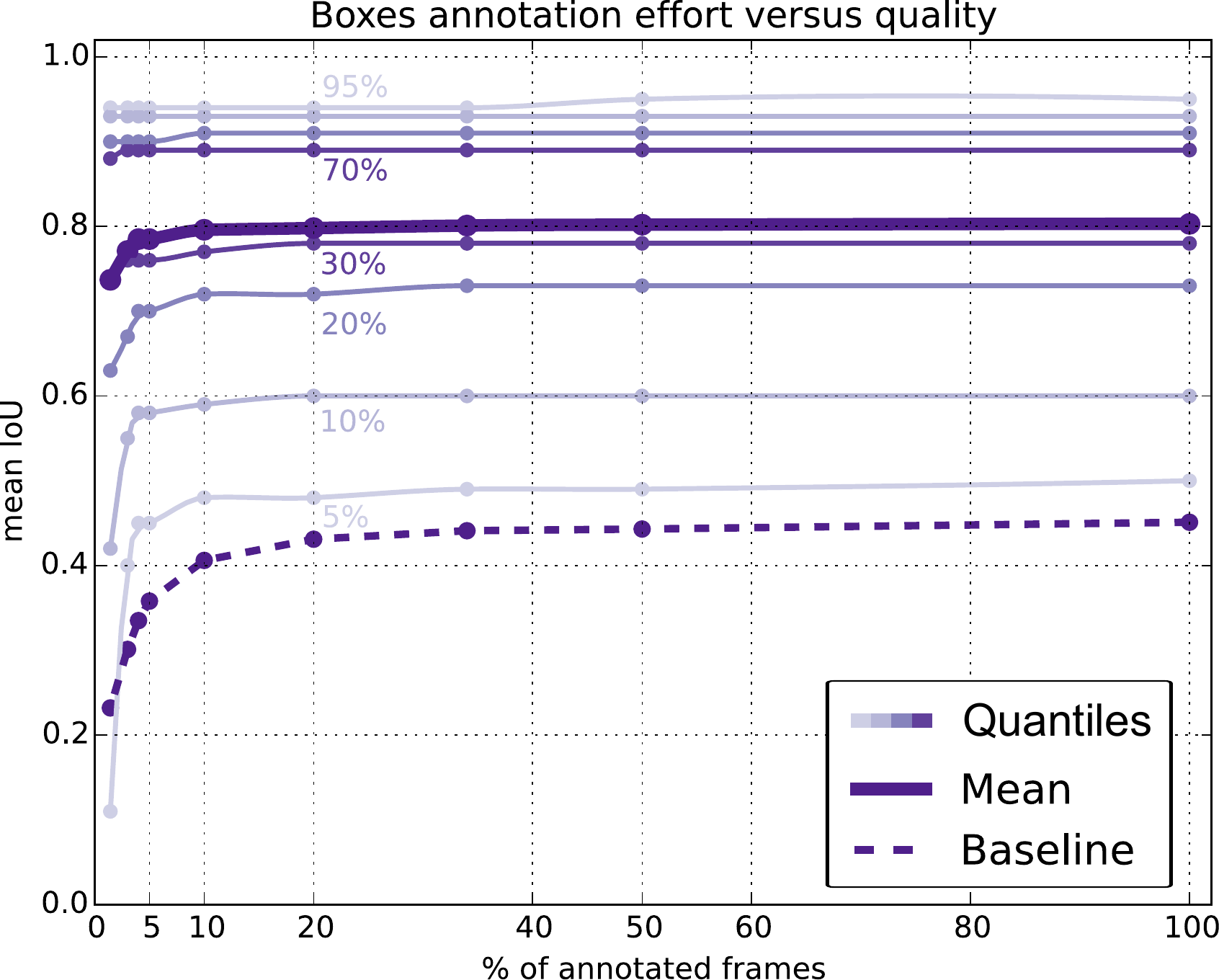}
\par\end{centering}
}\hspace*{\fill}
\caption{\label{fig:annotation-density}Percent of annotated frames versus
video object segmentation quality. We report mean IoU, and quantiles
at $5,\,10,\,20,\,30,\,70,\,80,\,90,\,\text{and}\,95\%$. Result on
DAVIS dataset, using segment or box annotations. The baseline simply
copies the annotations to adjacent frames.
Discussion in \S\ref{sec:few-frames-results}.
}\vspace{-1em}
\end{figure}

In some applications, e.g. video editing for movie production, one may want to consider more than a single frame annotation on videos. Figure \ref{fig:annotation-density} shows the video object segmentation quality
result when considering different number of annotated frames, on the DAVIS dataset. We show results for both pixel accurate segmentation per frame, or bounding box annotation per frame.

For these experiments, we run our method twice, forward and backwards; and for each frame pick the result closest in time to the annotated frame (either from forward or backwards propagation). Here,
the online fine-tuning uses all annotated frames instead of only the first one. For the experiments with box annotations (Figure \ref{fig:box-annotations-density}), we use a similar procedure to $\mathtt{MaskTrack}_{Box}$.
Box annotations are first converted to segments, and then apply $\mathtt{MaskTrack}$ as-is, treating these as the original segment annotations.

The evaluation reports the mean IoU results when annotating one frame only (same as table \ref{tab:VOS-results}), and every 40th, 30th, 20th, 10th, 5th, 3rd, and 2nd frame. Since DAVIS videos have length $\sim\negmedspace100\
\text{frames}$, 1 annotated frame corresponds to $\sim\negmedspace1\%$, otherwise annotations every 20th is $5\%$ of annotated frames, 10th $10\%$, 5th $20\%$, etc. We follow the same DAVIS evaluation protocol as
\S\ref{sec:single-frame-results}, ignoring first and last frame, and choosing to include the annotated frames in the evaluation (this is particularly relevant for the box annotation results).

Other than mean IoU we also show the quantile curves indicating the cutting line for the $5\%,\,10\%,\,20\%,\, \text{etc.}$ lowest quality video frame results. This gives a hint of how much targeted additional annotations might be needed.
The higher mean IoU of these quantiles are, the better.

The baseline for these experiments consists in directly copying the ground truth annotations from the nearest annotated neighbour. This baseline indicates "level zero" for our results. For visual clarity, we only include the mean value
for the baseline.

\paragraph{Analysis}
We can see that Figures \ref{fig:segment-annotations-density} and \ref{fig:box-annotations-density} show slightly different trends. When using segment annotations (Figure \ref{fig:segment-annotations-density}),
the baseline quality increases
steadily until reaching IoU 1 when all frames are annotated. Our $\mathtt{MaskTrack}$  approach provides large gains with $30\%$ of annotated frames or less. For instance when annotating $10\%$ of the frames we reach mIoU $0.86$,
notice that also the $20\%$ quantile is at $0.81\ \text{mIoU}$. This means with only $10\%$ of annotated frames, $80\%$ of all video frames will have a mean IoU above $0.8$, which is good enough to be used for many applications
, or can serve as initialization for a refinement process. With $10\%$ of annotated frames the baseline only reaches $0.64\ \text{mIoU}$.
When using box annotations (Figure \ref{fig:box-annotations-density}) the quality of the baseline and our method saturates. There is only so much information our instance segmenter can estimate from boxes. After $10\%$ of
annotated frames, not much additional gain is obtained. Interestingly, the mean IoU and $30\%$ quantile here both reach $\sim\negmedspace0.8\ \text{mIoU}$ range. Additionally, $70\%$ of the frames have IoU above $0.89$.

\paragraph{Conclusion}
Results indicate that with only $10\%$ of annotated frames we can reach satisfactory quality, even when using only bounding box annotations. We see that with moving from one annotation per video to two or three
frames ($1\%\negmedspace\rightarrow 3\%\negmedspace\rightarrow\negmedspace4\%$) quality increases sharply, showing that our system can adequately leverage a few extra annotations per video.

\section{Conclusion}
\label{sec:conclusion}
%
%
%

We have presented a novel approach to video object segmentation. By treating video object segmentation as a guided instance segmentation problem,
we have proposed to use a pixel labelling convnet for frame-by-frame segmentation.
By exploiting both offline and online training with image annotations only our approach is able to produce highly accurate video object segmentation.
The proposed system is generic and reaches competitive performance on three extremely heterogeneous video segmentation benchmarks, using the same model and parameters across all videos.
The method can handle different types of input annotations and our results are competitive even when using only bounding box annotations (instead of segmentation masks).

We provided a detailed ablation study, and explored the effect of varying the amount of annotations per video.
Our results show that with only one annotation every 10th frame we can reach $85\%$ mIoU quality.
Considering we only do per-frame instance segmentation without any form of global optimization,
we deem these results encouraging to achieve high quality via additional post-processing.

We believe the use of labelling convnets for video object segmentation is a promising strategy. Future work should consider exploring more sophisticated network architectures, incorporating temporal dimension
and adding global optimization strategies.

\FloatBarrier
{\small
\bibliographystyle{ieee}
\bibliography{2017_cvpr_video_object_segmentation}
}

\renewcommand{\thetable}{S\arabic{table}}
\renewcommand{\thefigure}{S\arabic{figure}}

\newpage
\clearpage
\appendix
\part*{Supplementary material}
\section{Content}
\label{sec:content}

This supplementary material provides both additional
quantitative and qualitative results, as well as an attached
video.

\begin{itemize}

\item Section \ref{sec:add_res} provides additional quantitative results for DAVIS, YoutubeObjects, and SegTrackv-2 (see Tables \ref{tab:youtube_cat} - \ref{tab:segtrack_seq}).
\item Detailed attribute-based evaluation is reported in Section \ref{sec:attr_eval} and Table \ref{tab:attr-eval}.
\item The dataset specific tuning for additional ingredients is described in Section \ref{sec:dat_tun}.
\item Additional qualitative results with first frame box and segment supervision are presented in Section \ref{sec:add_qual_res} and Figure \ref{fig:qualitative-results}.
\item Examples of mask generation for the extra input channel are shown in Section \ref{sec:mask_gen} and Figure \ref{fig:offline-augmentation}.
\item Examples of optical flow magnitude images are presented in Section \ref{sec:flow_gen} and Figure \ref{fig:flow_images}.

%
%
%

\end{itemize}

\section{Additional quantitative results}
\label{sec:add_res}
In this section we present additional quantitative results for three different datasets: DAVIS \cite{Perazzi2016Cvpr}, YoutubeObjects \cite{Prest2012Cvpr}, and SegTrackv-2  \cite{Li2013Iccv}.
This section complements \S \ref{sec:single-frame-results} in the main paper.

\paragraph{DAVIS}

We present the per-sequence comparison with other state-of-the-art methods on DAVIS in Table \ref{tab:davis_seq}.

\begin{table}
\begin{centering}
\begingroup
\begin{tabular}{l|@{   }c@{   }c@{   }c@{   }c@{    }}
\multirow{2}{*}{Sequence} & \multicolumn{4}{c}{Method, mIoU}\\
& $\mathtt{BVS}$ \cite{Maerki2016Cvpr} & \ $\mathtt{ObjFlow}$\cite{Tsai2016Cvpr} & \ $\mathtt{MaskTrack}$ & \  $\mathtt{\mathsmaller{+}Flow\mathsmaller{+}CRF}$ \tabularnewline
\hline
\hline
bear                & \bf 95.5    &     94.6    &     92.8  &		93.1 \\
blackswan           &     94.3    & \bf 94.7    &     91.9  &		90.3 \\
bmx-bumps           &     43.4    &     48.0    &     39.6  & \bf 57.1 \\
bmx-trees           & 38.2    &     14.9    &     32.1  &	\bf 	57.5 \\
boat                &     64.4    & \bf 80.8    &     78.2  &		54.7 \\
breakdance          &     50.0    &     49.6    &  59.4  &	\bf 	76.1 \\
breakdance-flare    &     72.7    &     76.5    & \bf 89.2  &		77.6 \\
bus                 &  86.3    &     68.2    &     79.1  &	\bf 	89.0 \\
camel               &     66.9    & \bf 86.7    &     80.4  &		80.1 \\
car-roundabout      &     85.1    &  90.0    &     82.8  &	\bf 	96.0 \\
car-shadow          &     57.8    &     84.6    &  90.3  &	\bf 	93.5 \\
car-turn            &     84.4    &     87.6    & \bf 92.3  &		88.6 \\
cows                &     89.5    &     91.0    & \bf 91.9  &		88.2 \\
dance-jump          &     74.5    & \bf 80.4    &     66.2  &		78.8 \\
dance-twirl         &     49.2    &     56.7    &  67.8  &	\bf 	84.4 \\
dog                 &     72.3    & 89.7    &     86.8  &	\bf 	90.8 \\
dog-agility         &     34.5    & \bf 86.0    &     83.7  &		78.9 \\
drift-chicane       &     3.3     &  17.5    &     0.5   &	\bf 	86.2 \\
drift-straight      &     40.2    &     31.4    &  46.0  &	\bf 	56.0 \\
drift-turn          &     29.9    &     3.5     & \bf 86.5  &		86.0 \\
elephant            &     84.9    &     87.9    & \bf 91.4  &		87.2 \\
flamingo            & \bf 88.1    &     87.3    &     70.9  &		79.0 \\
goat                &     66.1    & \bf 86.5    &     85.8  &		84.5 \\
hike                &     75.5    & \bf 93.4    &     74.5  &		93.1 \\
hockey              &     82.9    & \bf 85.0    &     84.0  &		83.4 \\
horsejump-high      &     80.1    & \bf 86.2    &     78.4  &		81.8 \\
horsejump-low       &     60.1    & \bf 82.2    &     79.6  &		80.6 \\
kite-surf           &     42.5    & \bf 70.2    &     58.7  &		60.0 \\
kite-walk           & \bf 87.0    &     85.0    &     77.4  &		64.5 \\
libby               &     77.6    &     59.4    & \bf 78.8  &		77.5 \\
lucia               &  90.1    &     89.7    &     88.4  &	\bf 	91.1 \\
mallard-fly         & \bf 60.6    &     55.0    &     56.7  &		57.2 \\
mallard-water       &     90.7    &     89.9    & \bf 91.0  &		90.4 \\
motocross-bumps     &     40.1    &     48.5    &  53.9  &	\bf 	59.9 \\
motocross-jump      &     34.1    &     59.4    & \bf 69.0  &		68.3 \\
motorbike           &  56.3    &     47.8    &     46.5  &	\bf 	56.7 \\
paragliding         &     87.5    &  94.7    &     93.2  &	\bf 	95.9 \\
paragliding-launch  & \bf 64.0    &     63.7    &     58.9  &		62.1 \\
parkour             &     75.6    & \bf 86.1    &     85.3  &		88.2 \\
rhino               &     78.2    &     89.5    & \bf 93.2  &		91.1 \\
rollerblade         &     58.8    & \bf 88.6    &     33.0  &		78.7 \\
scooter-black       &     33.7    &  76.5    &     64.9  &	\bf 	82.4 \\
scooter-gray        &     50.8    &     29.6    &  81.7  &	\bf 	82.9 \\
soapbox             &     78.9    &     68.9    &  86.1  &	\bf 	89.9 \\
soccerball          &     84.4    &     8.0     &  85.8  &	\bf 	89.0 \\
stroller            &     76.7    & \bf 87.7    &     86.2  &		85.4 \\
surf                &     49.2    & \bf 95.6    &     92.6  &		92.8 \\
swing               &     78.4    &     60.4    &  80.7  &	\bf 	81.9 \\
tennis              &     73.7    &     81.8    & \bf 87.3  &		86.2 \\
train               &     87.2    & \bf 91.7    &     90.8  &		90.4 \\
\hline
Mean                &  66.5   &    71.1   &   74.8       &	 \bf	80.3
\end{tabular}\endgroup
\par\end{centering}
\caption{\label{tab:davis_seq}Per-sequence results on the DAVIS dataset.}
\end{table}

\paragraph{SegTrack-v2}

Table \ref{tab:segtrack_seq} reports the per-sequence comparison with other state-of-the-art methods on SegTrack-v2.

\begin{table}[h]
\begin{centering}
\begingroup
\begin{tabular}{l@{  }|@{   }c@{   }c@{   }c@{   }c@{   }}
\multirow{2}{*}{Sequence} & \multicolumn{4}{c}{Method, mIoU}\\
& $\mathtt{BVS}$ \cite{Maerki2016Cvpr} & $\mathtt{ObjFlow}$ \cite{Tsai2016Cvpr}& $\mathtt{TRS}$ \cite{Xiao2016Cvpr} & $\mathtt{MaskTrack}$ \tabularnewline
\hline
\hline
bird of paradise & 89.7 & 87.1 & \bf 90.0 &84.0 \\
birdfall & 65.3 & 52.9 &  \bf 72.5& 56.6 \\
bmx\#1 & 67.1 &  \bf 87.9 & 86.1 &81.9 \\
bmx\#2 & 3.2 & 4.0 & \bf  40.3 & 0.1 \\
cheetah\#1 & 5.4 & 25.9 & 61.2 &  \bf 69.3 \\
cheetah\#2 & 9.2 & 37.2 &  \bf 39.4 & 17.4 \\
drift\#1 & 68.5 & \bf  77.9 & 70.7 & 47.4 \\
drift\#2 & 32.7 & 27.4 & 70.7 &  \bf 70.9 \\
frog & 76.1 & 78.4 & 80.2 &  \bf  85.3 \\
girl & 86.5 & 84.2 & 86.4 &  \bf 86.8 \\
hummingbird\#1 & 53.2 & \bf  67.2 & 53.0 & 39.0 \\
hummingbird\#2 & 28.7 & 68.5 &  \bf 70.5 & 49.6 \\
monkey & 85.7 & 87.8 & 83.1 &  \bf 89.3 \\
monkeydog\#1 & 40.5 & 47.1 &  \bf 74.0 & 25.3 \\
monkeydog\#2 & 17.1 & 21.0 &  \bf 39.6 & 31.7 \\
parachute & 93.7 & 93.3 &  \bf 95.9 & 93.7 \\
penguin\#1 & 81.6 & 80.4 & 53.2 &  \bf 93.7 \\
penguin\#2 & 82.0 & 83.5 & 72.9 &  \bf 85.2 \\
penguin\#3 & 78.5 & 83.9 & 74.4 &  \bf 90.1 \\
penguin\#4 & 76.4 & 86.2 & 57.2 & \bf  90.5 \\
penguin\#5 & 47.8 &  \bf 82.3 & 63.5 & 78.4 \\
penguin\#6 & 84.3 & 87.3 & 65.7 &  \bf 89.3 \\
soldier & 55.3 & \bf  86.8 & 76.3 & 82.0 \\
worm & 65.4 & 83.2 & \bf  82.4 &80.4 \\
\hline
Mean  &  58.4   &     67.5  & \bf 69.1 & 67.4   \\
\end{tabular}\endgroup
\par\end{centering}
\caption{\label{tab:segtrack_seq}Per-sequence results on the SegTrack-v2 dataset.}
\end{table}

\paragraph{YoutubeObjects}
The per-category comparison with other state-of-the-art methods on YoutubeObjects is shown in Table \ref{tab:youtube_cat}.

\section{Attribute-based evaluation}
\label{sec:attr_eval}

Table \ref{tab:attr-eval} presents a more detailed evaluation on DAVIS using video attributes and complements Figure \ref{fig:attribute_analysis} in the main paper.

The attribute based evaluation shows that our generic model, $\mathtt{MaskTrack}$,
is robust to various video challenges present in DAVIS. It compares favourably on any subset of videos sharing the same attribute, except camera-shake, where $\mathtt{ObjFlow}$ \cite{Tsai2016Cvpr} marginally outperforms our approach.

We observe that $\mathtt{MaskTrack}$ handles well fast-motion, appearance change and out-of-view, where competitive methods are failing \cite{Maerki2016Cvpr, Tsai2016Cvpr}.\\
Furthermore, incorporating optical flow information and CRF post-processing into $\mathtt{MaskTrack}$ substantially increases robustness on all categories, reaching over 70$\%$ mIoU on each subcategory.
In particular, $\mathtt{MaskTrack\mathsmaller{+}Flow \mathsmaller{+}CRF}$
better discriminates cases of low resolution, scale-variation and appearance change.

\begin{table}
\begin{centering}
\begingroup
\begin{tabular}{l@{   }|@{   }c@{   }c@{   }c@{   }}
\multirow{2}{*}{Category} & \multicolumn{3}{c}{Method, mIoU}\\
& $\mathtt{BVS}$ \cite{Maerki2016Cvpr} & $\mathtt{ObjFlow}$ \cite{Tsai2016Cvpr}& $\mathtt{MaskTrack}$ \tabularnewline
\hline
\hline
aeroplane& 80.8& \textbf{85.3} & 81.6\tabularnewline
bird & 76.4 & \textbf{83.1}& 82.9 \tabularnewline
boat& 60.1 & 70.6 & \textbf{74.7} \tabularnewline
car& 56.7 & \textbf{68.8} & 66.9 \tabularnewline
cat& 52.7 & 60.6 & \textbf{69.6}\tabularnewline
cow& 64.8& 71.5 & \textbf{75.0} \tabularnewline
dog& 61.6 & 71.6 & \textbf{75.2} \tabularnewline
horse& 53.1 & 62.3 & \textbf{64.9} \tabularnewline
motorbike& 41.6 & \textbf{59.9} & 49.8 \tabularnewline
train& 62.1& 74.7 & \textbf{77.7} \tabularnewline
\hline
Mean per object& 59.7 & 70.1 & \textbf{71.7} \tabularnewline
Mean per class& 61.0 & 70.9 & \textbf{71.9} \tabularnewline
\end{tabular}\endgroup
\par\end{centering}
\caption{\label{tab:youtube_cat}Per-category results on the YoutubeObjects dataset.}
\end{table}

\begin{table*}
\begin{centering}
\begingroup
\begin{tabular}{l|ccc||cc}
\multirow{2}{*}{Attribute} & \multicolumn{5}{c}{Method, mIoU}\\
& $\mathtt{BVS}$ \cite{Maerki2016Cvpr} & $\mathtt{ObjFlow}$ \cite{Tsai2016Cvpr}& $\mathtt{MaskTrack}$ & $\mathtt{MaskTrack\mathsmaller{+}Flow}$ & $\mathtt{MaskTrack\mathsmaller{+}Flow \mathsmaller{+}CRF}$\tabularnewline
\hline
\hline
Appearance change & 0.46& 0.54& \textbf{0.65} & 0.75& \textbf{0.76}\tabularnewline
Background clutter& 0.63 & 0.68 & \textbf{0.77} & 0.78 & \textbf{0.79}\tabularnewline
Camera-shake& 0.62 & \textbf{0.72} & 0.71 & 0.77 & \textbf{0.78}\tabularnewline
Deformation& 0.7& 0.77& \textbf{0.77} &0.78 & \textbf{0.8}\tabularnewline
Dynamic background& 0.6 & 0.67 & \textbf{0.69}& 0.75 & \textbf{0.76}\tabularnewline
Edge ambiguity& 0.58 & 0.65& \textbf{0.68} & \textbf{0.74} & \textbf{0.74}\tabularnewline
Fast-motion & 0.53 & 0.55& \textbf{0.66} & 0.74 & \textbf{0.75}\tabularnewline
Heterogeneous object& 0.63 & 0.66 & \textbf{0.71} & 0.77 &\textbf{0.79}\tabularnewline
Interacting objects& 0.63 & 0.68 & \textbf{0.74} & 0.75 &\textbf{0.77}\tabularnewline
Low resolution& 0.59& 0.58& \textbf{0.6} & 0.75 & \textbf{0.77}\tabularnewline
Motion blur& 0.58& 0.6& \textbf{0.66}& 0.72 & \textbf{0.74}\tabularnewline
Occlusion& 0.68 & 0.66 & \textbf{0.74} & 0.75 & \textbf{0.77}\tabularnewline
Out-of-view& 0.43 & 0.53 & \textbf{0.66} & \textbf{0.71} & \textbf{0.71}\tabularnewline
Scale variation& 0.49 & 0.56 & \textbf{0.62}& 0.72 & \textbf{0.73}\tabularnewline
Shape complexity& 0.67 & 0.69 & \textbf{0.71} & 0.72 & \textbf{0.75}\tabularnewline
\end{tabular}\endgroup
\par\end{centering}
\caption{\label{tab:attr-eval}Attribute based evaluation on DAVIS.}
\end{table*}

\begin{figure*}
\begin{centering}
\setlength{\tabcolsep}{1pt}
\begin{tabular}{ccccccc}

\begin{turn}{90}
{\footnotesize{\hspace{1em} Box}}
\end{turn} &
\includegraphics[width=0.16\textwidth,height=0.06\textheight]{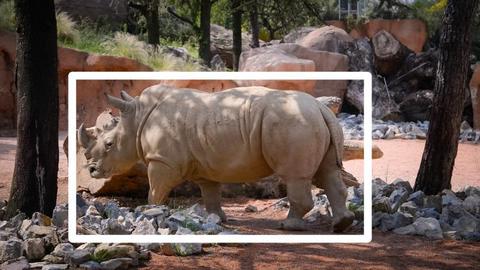} &
\includegraphics[width=0.16\textwidth,height=0.06\textheight]{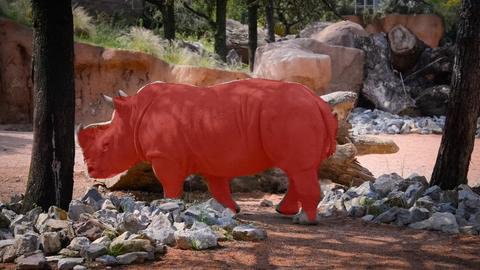} &
\includegraphics[width=0.16\textwidth,height=0.06\textheight]{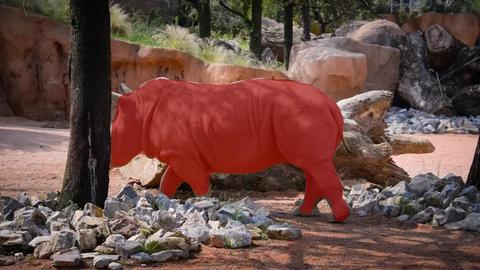} &
\includegraphics[width=0.16\textwidth,height=0.06\textheight]{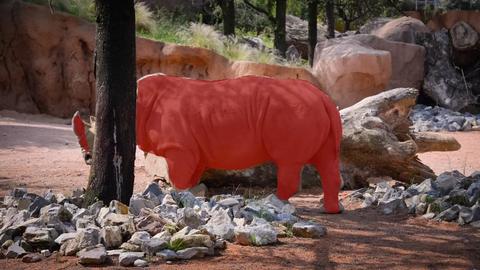} &
\includegraphics[width=0.16\textwidth,height=0.06\textheight]{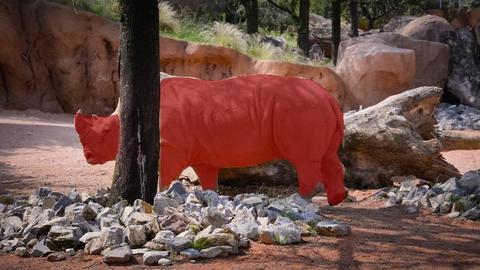} &
\includegraphics[width=0.16\textwidth,height=0.06\textheight]{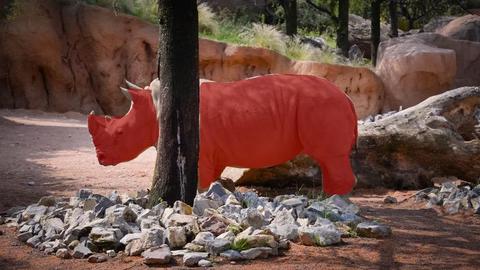} \tabularnewline
\begin{turn}{90}
{\footnotesize{\hspace{0.5em} Segment}}
\end{turn}  &
\includegraphics[width=0.16\textwidth,height=0.06\textheight]{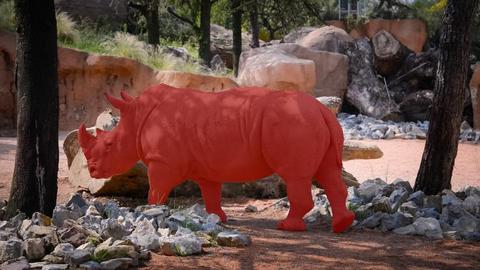} &
\includegraphics[width=0.16\textwidth,height=0.06\textheight]{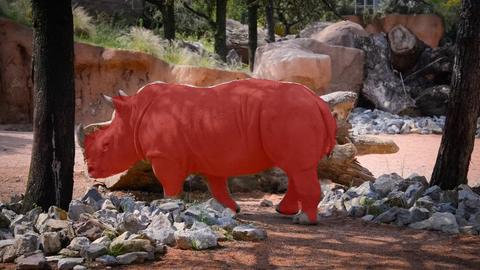} &
\includegraphics[width=0.16\textwidth,height=0.06\textheight]{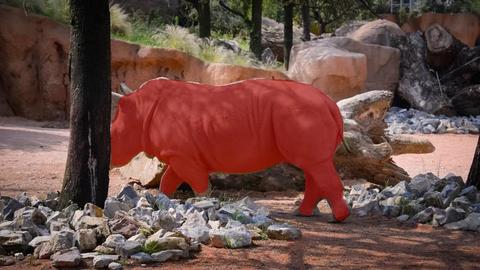} &
\includegraphics[width=0.16\textwidth,height=0.06\textheight]{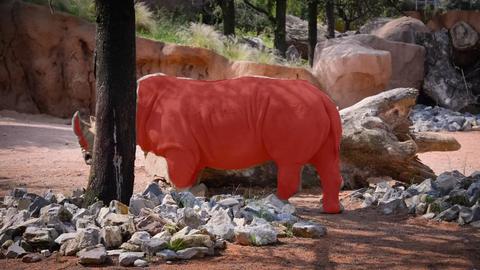} &
\includegraphics[width=0.16\textwidth,height=0.06\textheight]{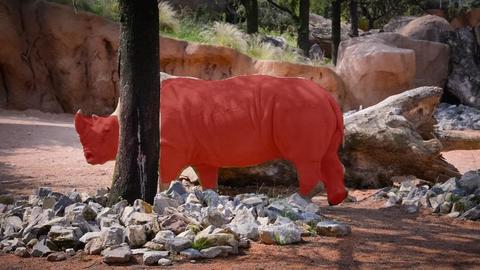} &
\includegraphics[width=0.16\textwidth,height=0.06\textheight]{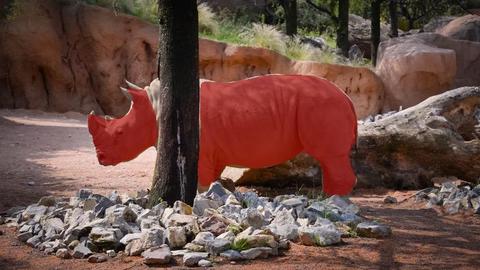} \tabularnewline
\\

\begin{turn}{90}
{\footnotesize{\hspace{1em} Box}}
\end{turn} &
\includegraphics[width=0.16\textwidth,height=0.06\textheight]{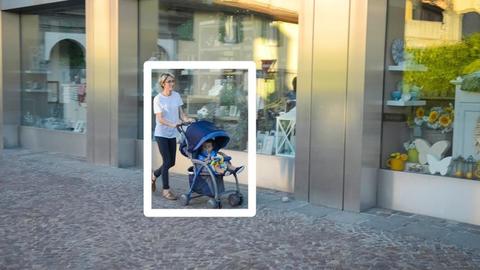} &
\includegraphics[width=0.16\textwidth,height=0.06\textheight]{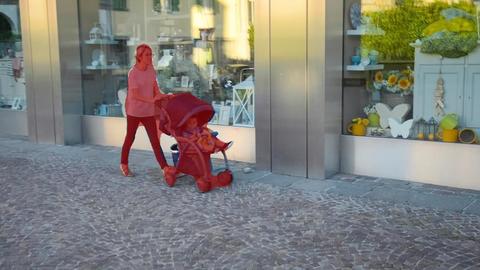} &
\includegraphics[width=0.16\textwidth,height=0.06\textheight]{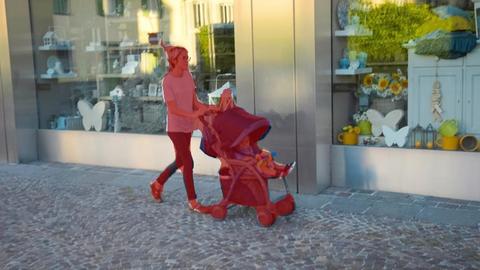} &
\includegraphics[width=0.16\textwidth,height=0.06\textheight]{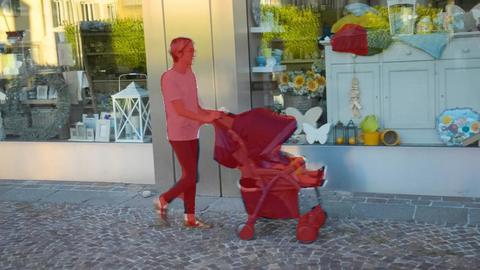} &
\includegraphics[width=0.16\textwidth,height=0.06\textheight]{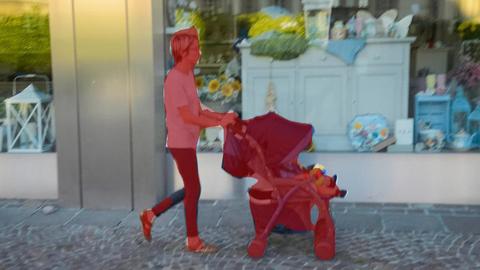} &
\includegraphics[width=0.16\textwidth,height=0.06\textheight]{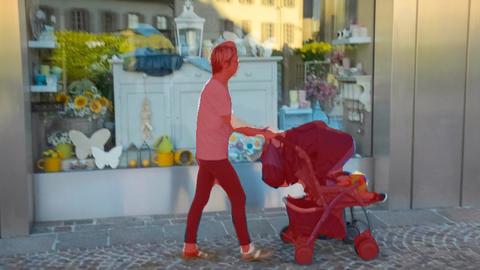} \tabularnewline
\begin{turn}{90}
{\footnotesize{\hspace{0.5em} Segment}}
\end{turn}  &
\includegraphics[width=0.16\textwidth,height=0.06\textheight]{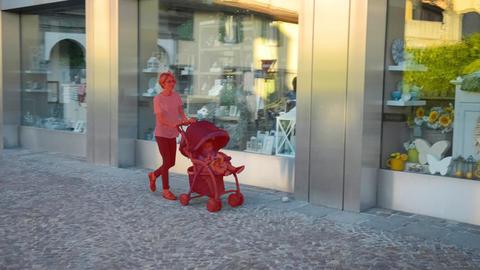} &
\includegraphics[width=0.16\textwidth,height=0.06\textheight]{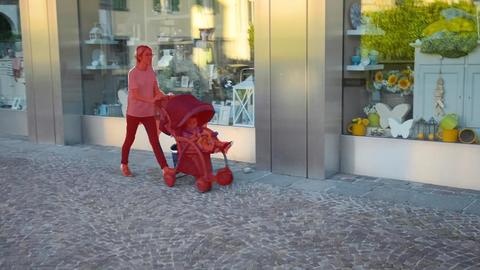} &
\includegraphics[width=0.16\textwidth,height=0.06\textheight]{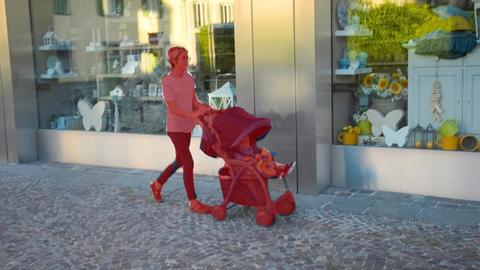} &
\includegraphics[width=0.16\textwidth,height=0.06\textheight]{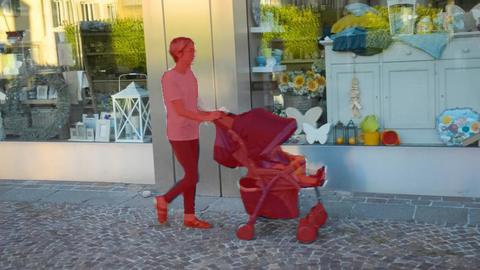} &
\includegraphics[width=0.16\textwidth,height=0.06\textheight]{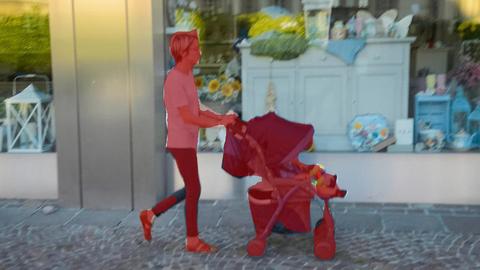} &
\includegraphics[width=0.16\textwidth,height=0.06\textheight]{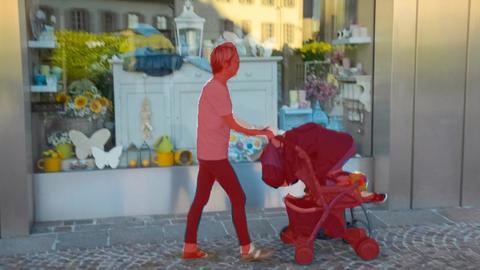} \tabularnewline

\\

\begin{turn}{90}
{\footnotesize{\hspace{1em} Box}}
\end{turn} &
\includegraphics[width=0.16\textwidth,height=0.06\textheight]{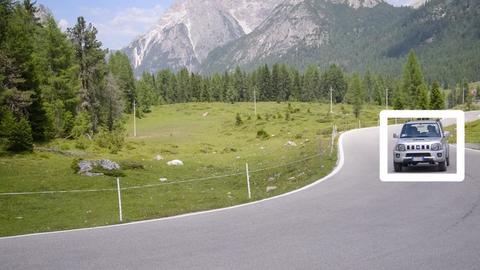} &
\includegraphics[width=0.16\textwidth,height=0.06\textheight]{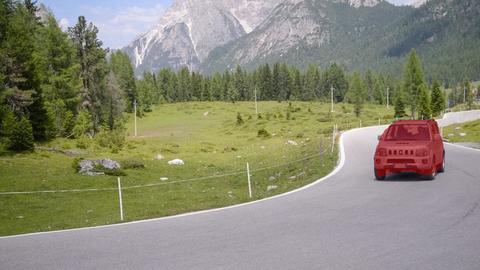} &
\includegraphics[width=0.16\textwidth,height=0.06\textheight]{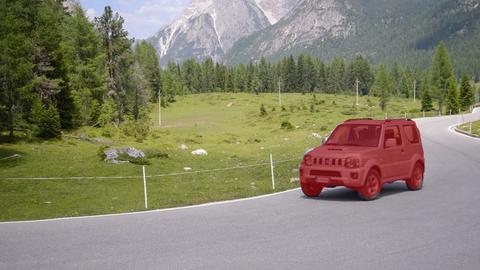} &
\includegraphics[width=0.16\textwidth,height=0.06\textheight]{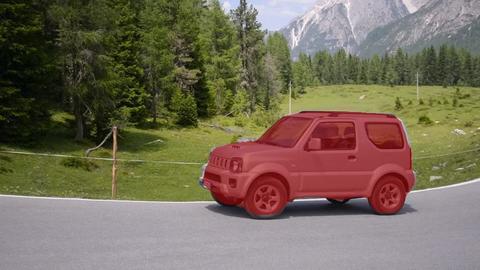} &
\includegraphics[width=0.16\textwidth,height=0.06\textheight]{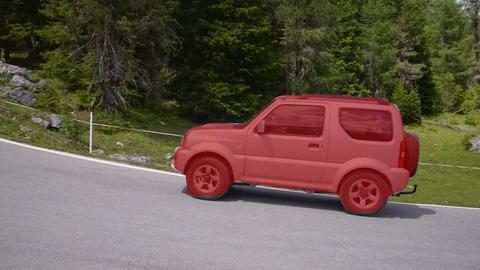} &
\includegraphics[width=0.16\textwidth,height=0.06\textheight]{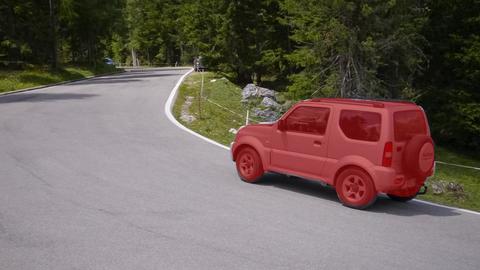} \tabularnewline
\begin{turn}{90}
{\footnotesize{\hspace{0.5em} Segment}}
\end{turn}  &
\includegraphics[width=0.16\textwidth,height=0.06\textheight]{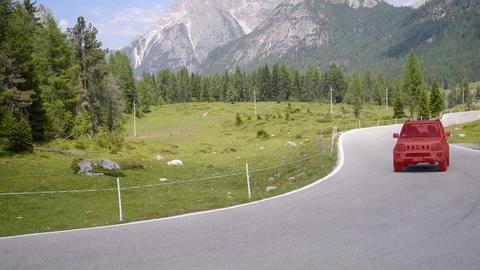} &
\includegraphics[width=0.16\textwidth,height=0.06\textheight]{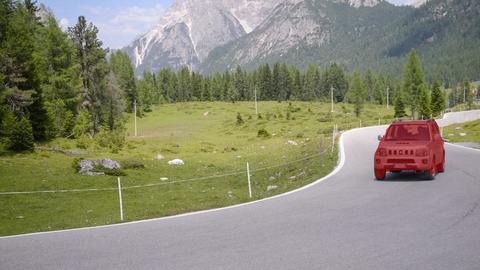} &
\includegraphics[width=0.16\textwidth,height=0.06\textheight]{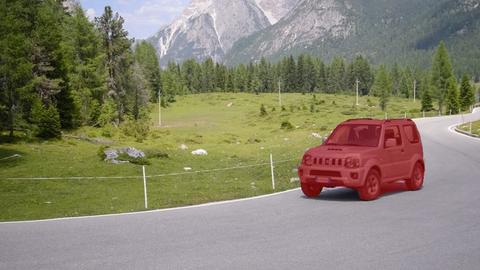} &
\includegraphics[width=0.16\textwidth,height=0.06\textheight]{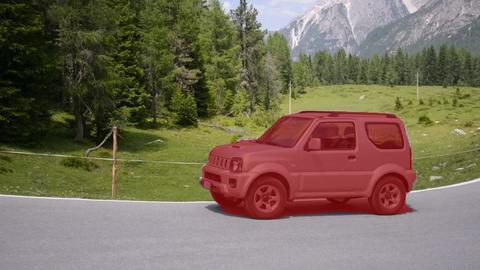} &
\includegraphics[width=0.16\textwidth,height=0.06\textheight]{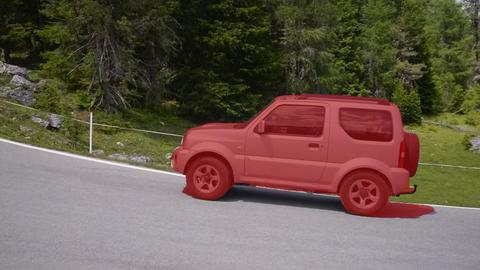} &
\includegraphics[width=0.16\textwidth,height=0.06\textheight]{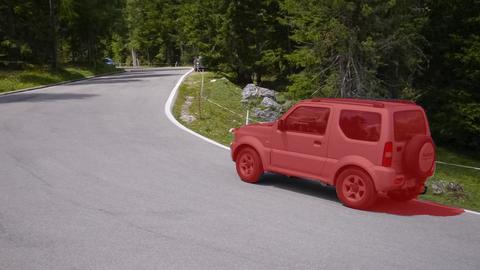} \tabularnewline

\\

\begin{turn}{90}
{\footnotesize{\hspace{1em} Box}}
\end{turn} &
\includegraphics[width=0.16\textwidth,height=0.06\textheight]{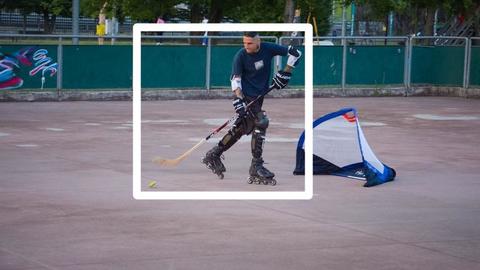} &
\includegraphics[width=0.16\textwidth,height=0.06\textheight]{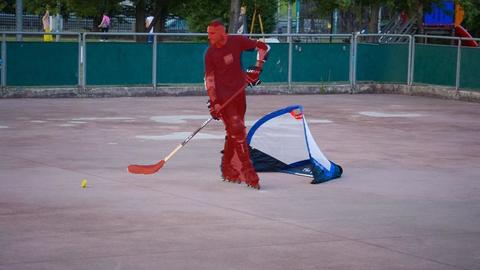} &
\includegraphics[width=0.16\textwidth,height=0.06\textheight]{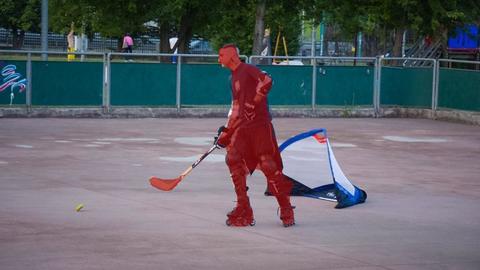} &
\includegraphics[width=0.16\textwidth,height=0.06\textheight]{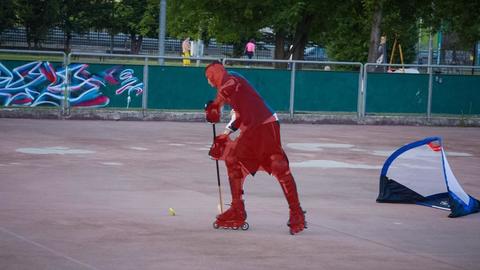} &
\includegraphics[width=0.16\textwidth,height=0.06\textheight]{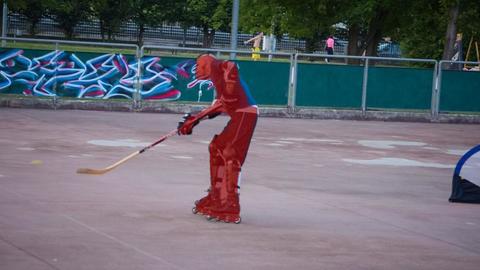} &
\includegraphics[width=0.16\textwidth,height=0.06\textheight]{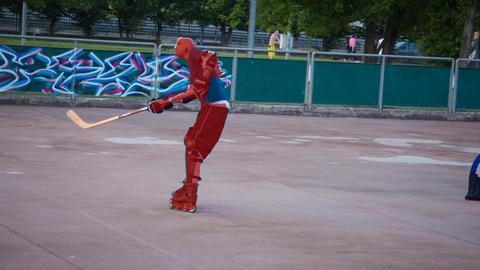} \tabularnewline
\begin{turn}{90}
{\footnotesize{\hspace{0.5em} Segment}}
\end{turn}  &
\includegraphics[width=0.16\textwidth,height=0.06\textheight]{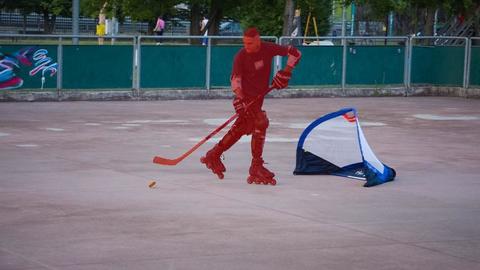} &
\includegraphics[width=0.16\textwidth,height=0.06\textheight]{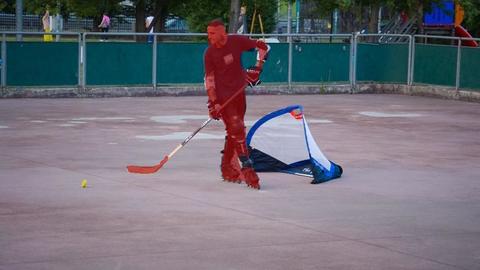} &
\includegraphics[width=0.16\textwidth,height=0.06\textheight]{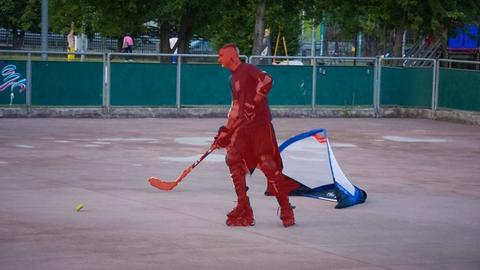} &
\includegraphics[width=0.16\textwidth,height=0.06\textheight]{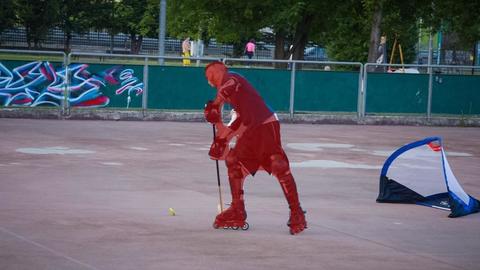} &
\includegraphics[width=0.16\textwidth,height=0.06\textheight]{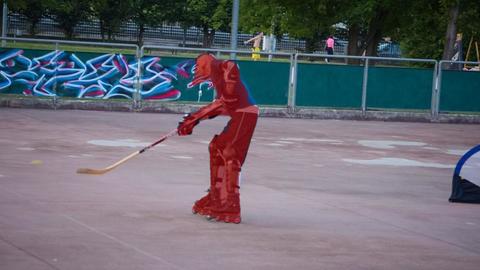} &
\includegraphics[width=0.16\textwidth,height=0.06\textheight]{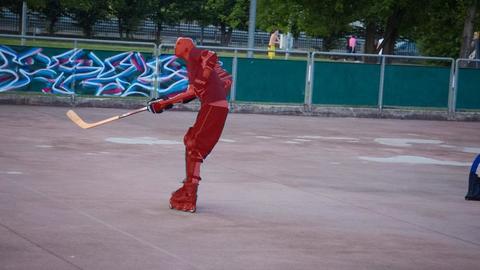} \tabularnewline

\\

\begin{turn}{90}
{\footnotesize{\hspace{1em} Box}}
\end{turn} &
\includegraphics[width=0.16\textwidth,height=0.06\textheight]{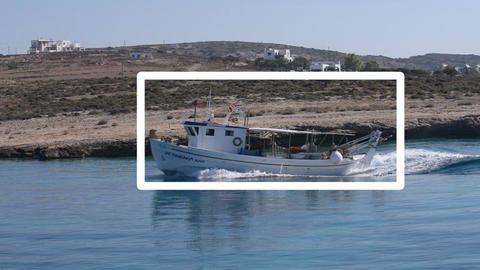} &
\includegraphics[width=0.16\textwidth,height=0.06\textheight]{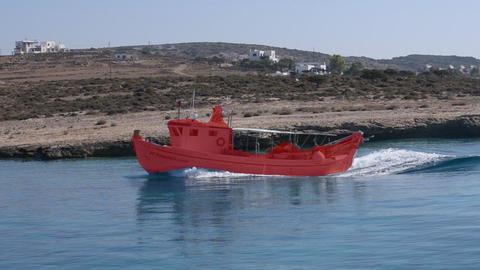} &
\includegraphics[width=0.16\textwidth,height=0.06\textheight]{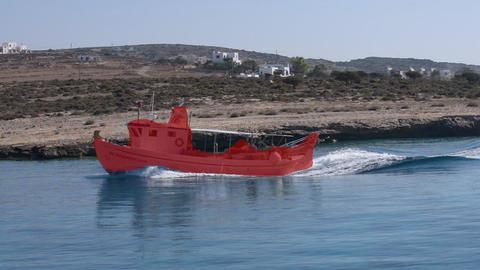} &
\includegraphics[width=0.16\textwidth,height=0.06\textheight]{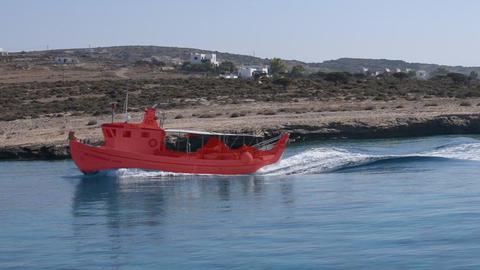} &
\includegraphics[width=0.16\textwidth,height=0.06\textheight]{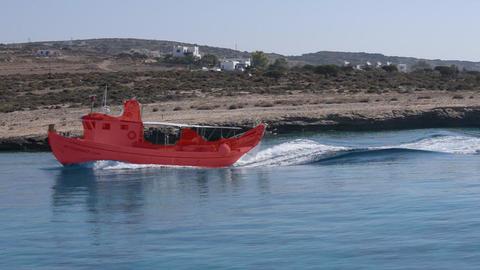} &
\includegraphics[width=0.16\textwidth,height=0.06\textheight]{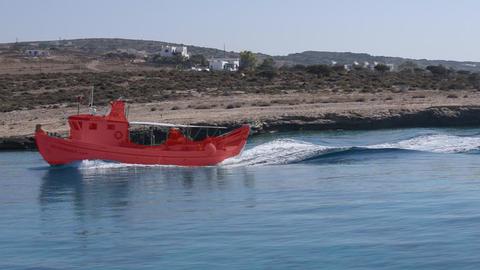} \tabularnewline
\begin{turn}{90}
{\footnotesize{\hspace{0.5em} Segment}}
\end{turn}  &
\includegraphics[width=0.16\textwidth,height=0.06\textheight]{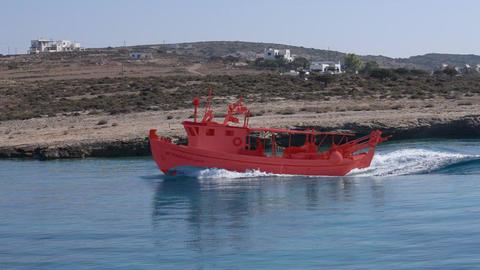} &
\includegraphics[width=0.16\textwidth,height=0.06\textheight]{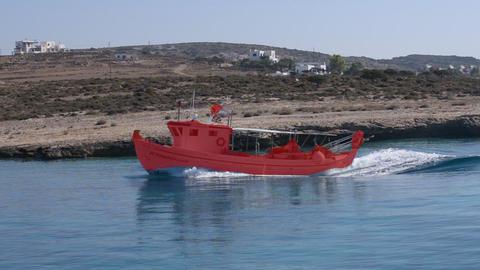} &
\includegraphics[width=0.16\textwidth,height=0.06\textheight]{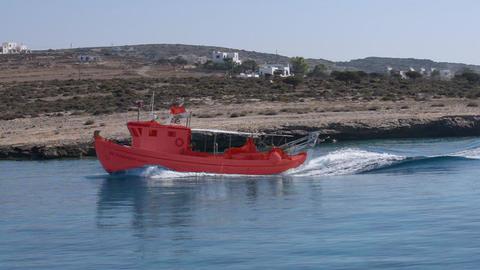} &
\includegraphics[width=0.16\textwidth,height=0.06\textheight]{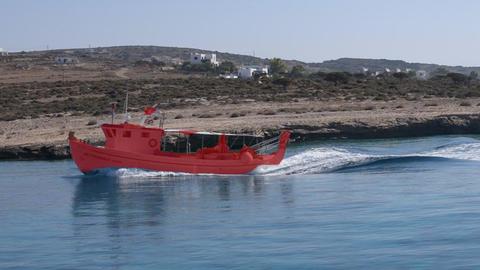} &
\includegraphics[width=0.16\textwidth,height=0.06\textheight]{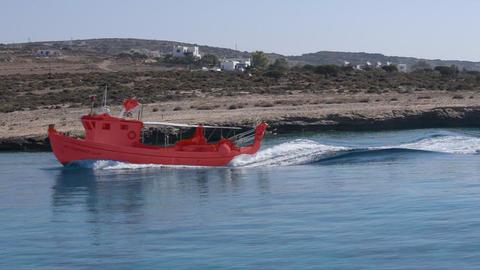} &
\includegraphics[width=0.16\textwidth,height=0.06\textheight]{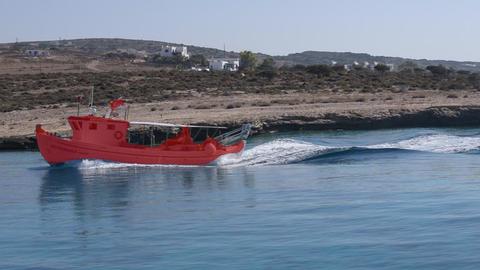} \tabularnewline

\\

\begin{turn}{90}
{\footnotesize{\hspace{1em} Box}}
\end{turn} &
\includegraphics[width=0.16\textwidth,height=0.06\textheight]{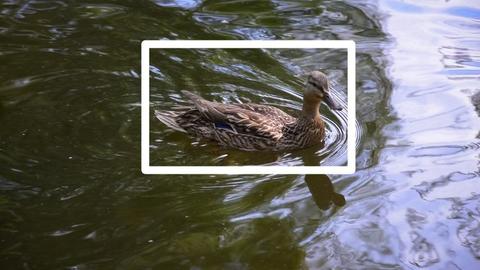} &
\includegraphics[width=0.16\textwidth,height=0.06\textheight]{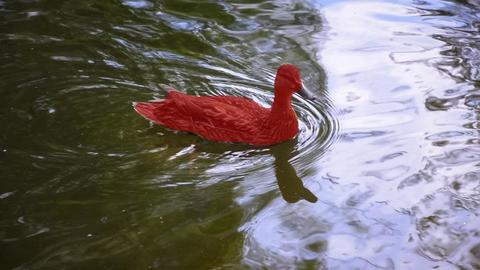} &
\includegraphics[width=0.16\textwidth,height=0.06\textheight]{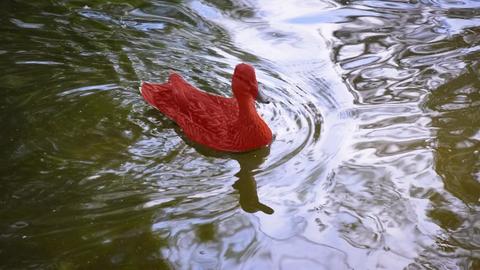} &
\includegraphics[width=0.16\textwidth,height=0.06\textheight]{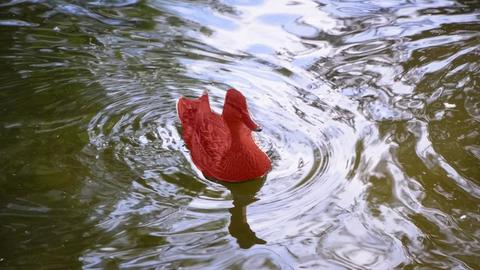} &
\includegraphics[width=0.16\textwidth,height=0.06\textheight]{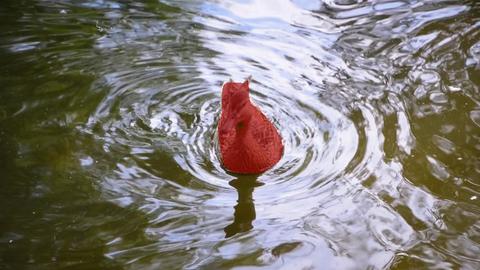} &
\includegraphics[width=0.16\textwidth,height=0.06\textheight]{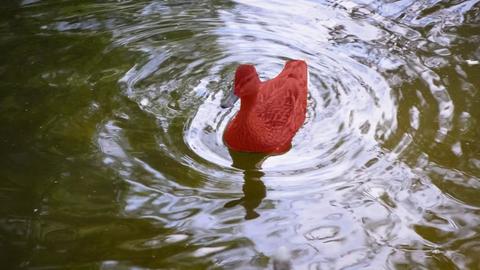} \tabularnewline
\begin{turn}{90}
{\footnotesize{\hspace{0.5em} Segment}}
\end{turn}  &
\includegraphics[width=0.16\textwidth,height=0.06\textheight]{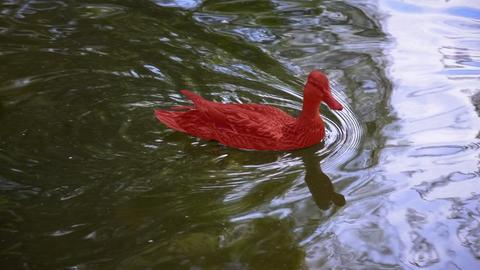} &
\includegraphics[width=0.16\textwidth,height=0.06\textheight]{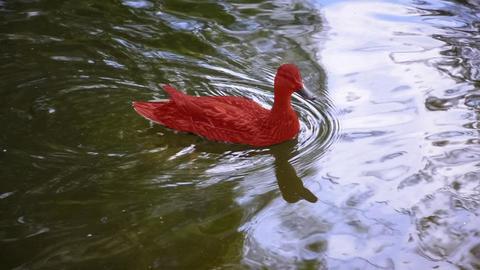} &
\includegraphics[width=0.16\textwidth,height=0.06\textheight]{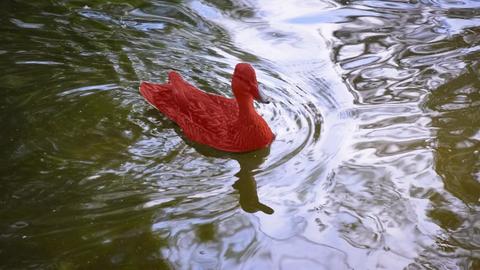} &
\includegraphics[width=0.16\textwidth,height=0.06\textheight]{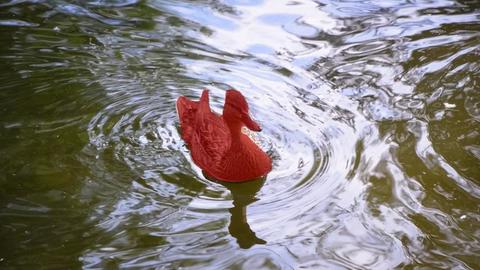} &
\includegraphics[width=0.16\textwidth,height=0.06\textheight]{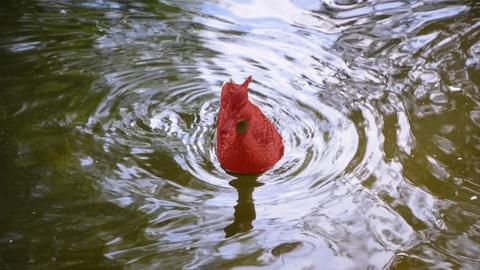} &
\includegraphics[width=0.16\textwidth,height=0.06\textheight]{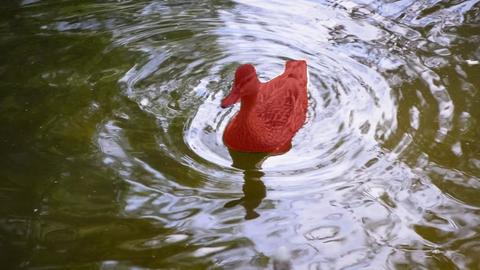} \tabularnewline

&\footnotesize{}1st frame annotation & \multicolumn{5}{c}{\footnotesize{} Results with $\mathtt{MaskTrack}_{Box}$ and $\mathtt{MaskTrack}$, the frames are chosen equally distant based on the video sequence length}
\end{tabular}
\par\end{centering}
\caption{\label{fig:qualitative-results2}
Qualitative results of  $\mathtt{MaskTrack}_{Box}$ and $\mathtt{MaskTrack}$ on Davis using 1st frame annotation supervision (box or segment).
By propagating annotation from the 1st frame, either from segment or just bounding box annotations, our system generates results comparable to ground truth.}
\end{figure*}

\begin{figure*}[t]
\begin{centering}
\begin{tabular}{@{}cc@{ }c@{ }c@{ }c@{ }c@{}}
\includegraphics[width=0.16\textwidth,height=0.08\textheight]{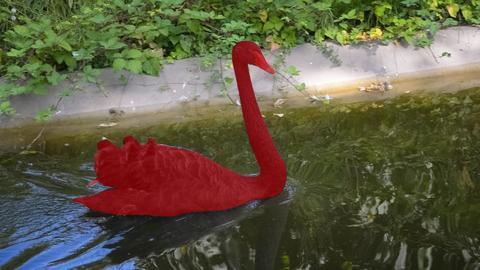} &
\includegraphics[width=0.16\textwidth,height=0.08\textheight]{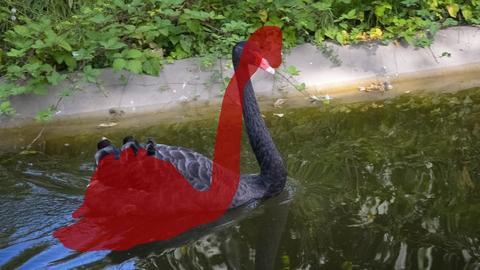} &
\includegraphics[width=0.16\textwidth,height=0.08\textheight]{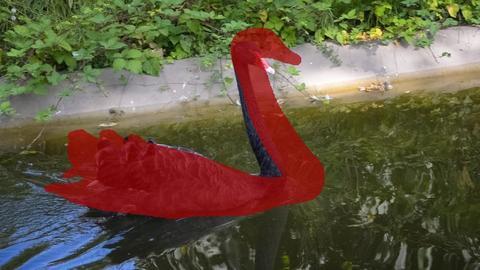} &
\includegraphics[width=0.16\textwidth,height=0.08\textheight]{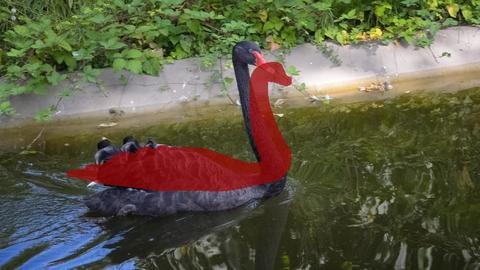} &
\includegraphics[width=0.16\textwidth,height=0.08\textheight]{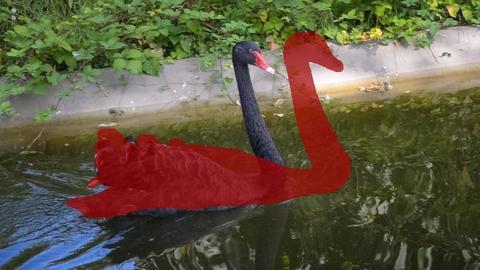} &
\includegraphics[width=0.16\textwidth,height=0.08\textheight]{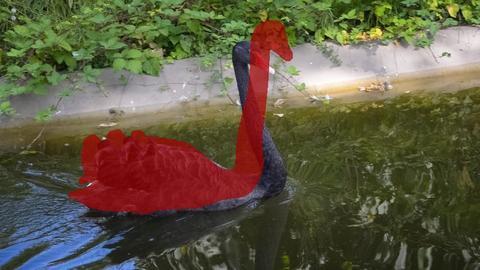} \\

\includegraphics[width=0.16\textwidth,height=0.08\textheight]{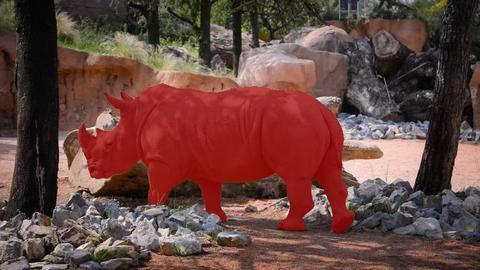} &
\includegraphics[width=0.16\textwidth,height=0.08\textheight]{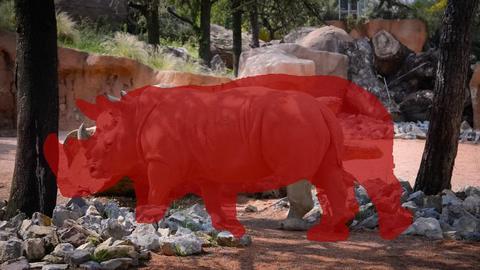} &
\includegraphics[width=0.16\textwidth,height=0.08\textheight]{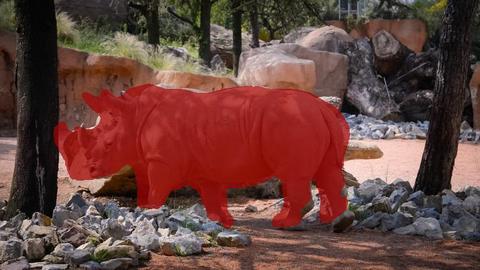} &
\includegraphics[width=0.16\textwidth,height=0.08\textheight]{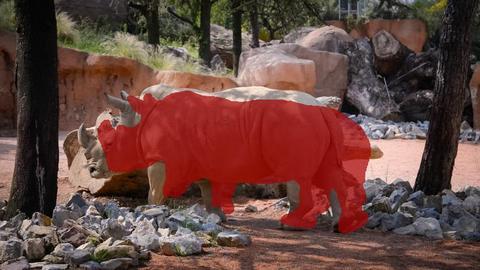} &
\includegraphics[width=0.16\textwidth,height=0.08\textheight]{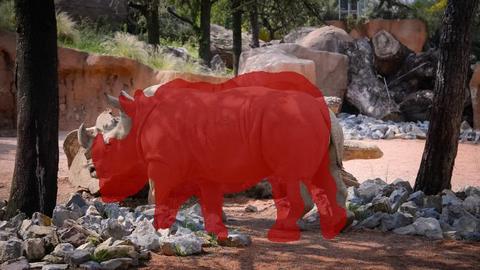} &
\includegraphics[width=0.16\textwidth,height=0.08\textheight]{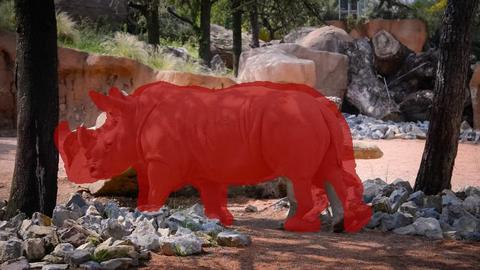} \\

\includegraphics[width=0.16\textwidth,height=0.08\textheight]{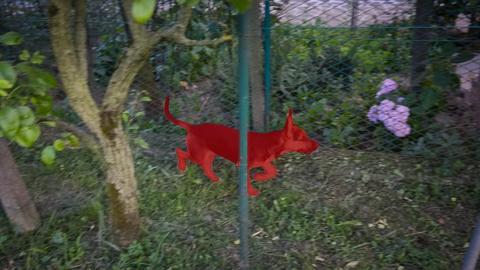} &
\includegraphics[width=0.16\textwidth,height=0.08\textheight]{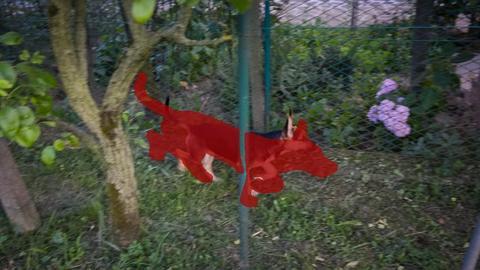} &
\includegraphics[width=0.16\textwidth,height=0.08\textheight]{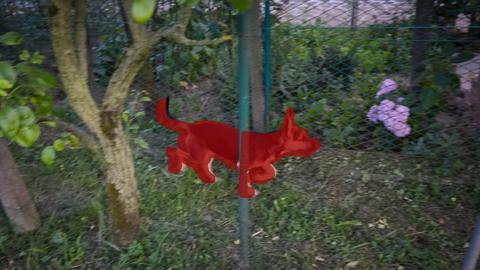} &
\includegraphics[width=0.16\textwidth,height=0.08\textheight]{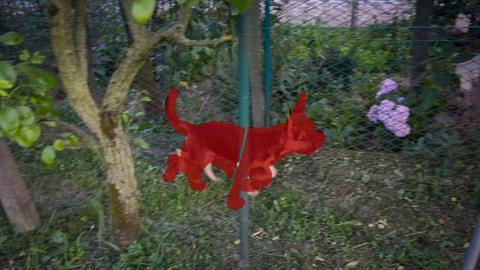} &
\includegraphics[width=0.16\textwidth,height=0.08\textheight]{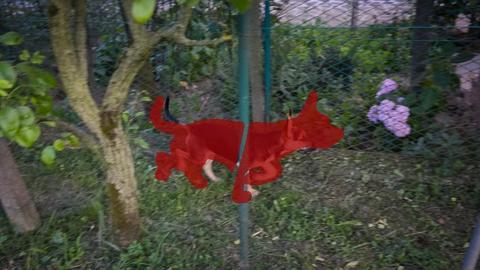} &
\includegraphics[width=0.16\textwidth,height=0.08\textheight]{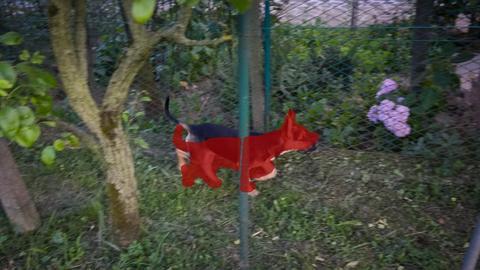} \\

\includegraphics[width=0.16\textwidth,height=0.08\textheight]{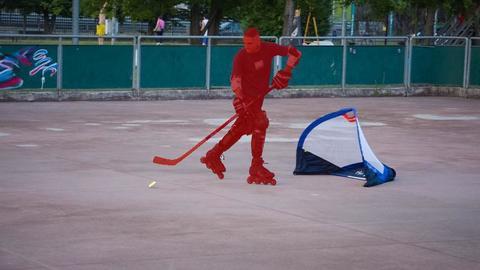} &
\includegraphics[width=0.16\textwidth,height=0.08\textheight]{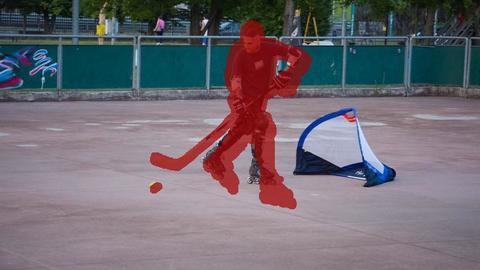} &
\includegraphics[width=0.16\textwidth,height=0.08\textheight]{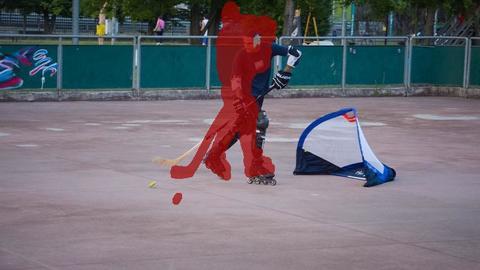} &
\includegraphics[width=0.16\textwidth,height=0.08\textheight]{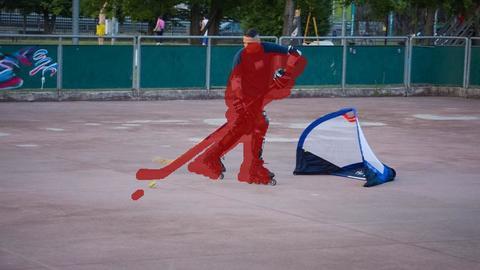} &
\includegraphics[width=0.16\textwidth,height=0.08\textheight]{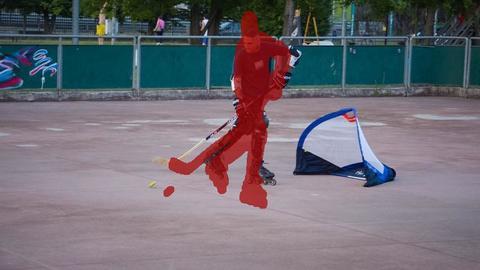} &
\includegraphics[width=0.16\textwidth,height=0.08\textheight]{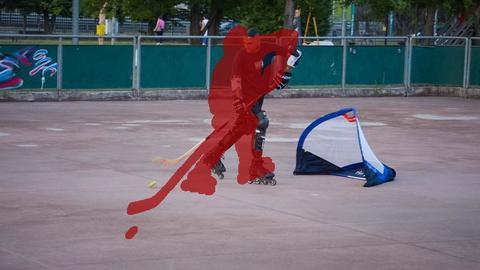} \\

\includegraphics[width=0.16\textwidth,height=0.08\textheight]{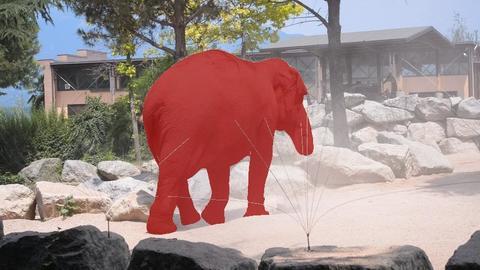} &
\includegraphics[width=0.16\textwidth,height=0.08\textheight]{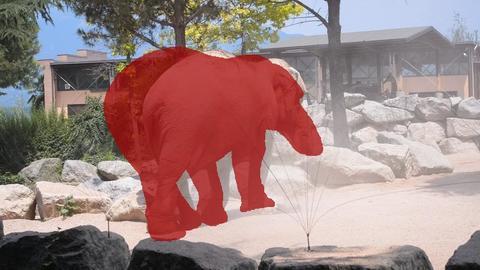} &
\includegraphics[width=0.16\textwidth,height=0.08\textheight]{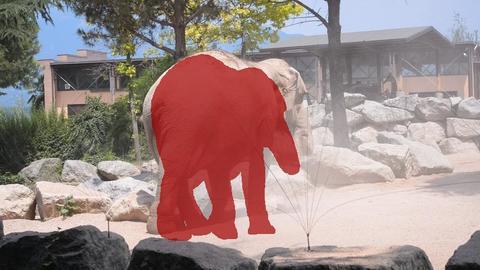} &
\includegraphics[width=0.16\textwidth,height=0.08\textheight]{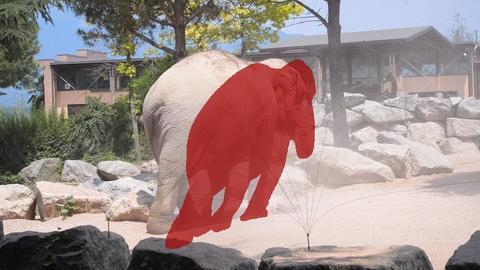} &
\includegraphics[width=0.16\textwidth,height=0.08\textheight]{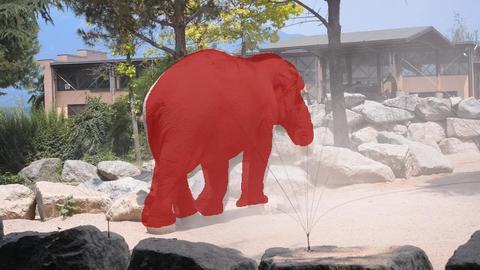} &
\includegraphics[width=0.16\textwidth,height=0.08\textheight]{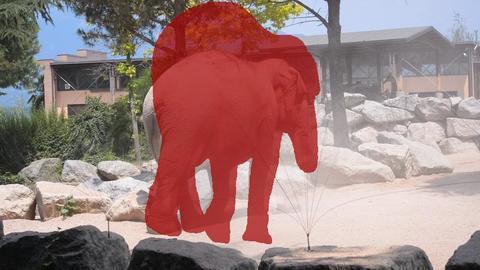} \\

\includegraphics[width=0.16\textwidth,height=0.08\textheight]{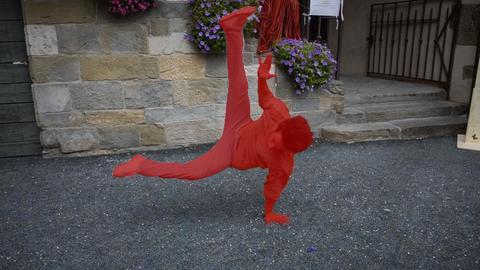} &
\includegraphics[width=0.16\textwidth,height=0.08\textheight]{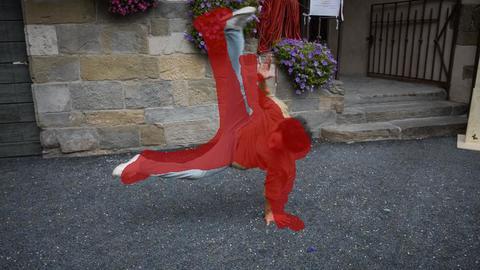} &
\includegraphics[width=0.16\textwidth,height=0.08\textheight]{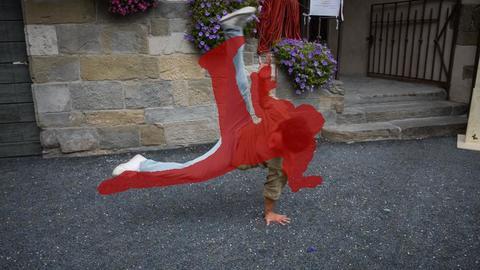} &
\includegraphics[width=0.16\textwidth,height=0.08\textheight]{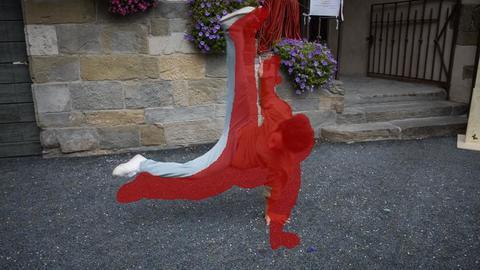} &
\includegraphics[width=0.16\textwidth,height=0.08\textheight]{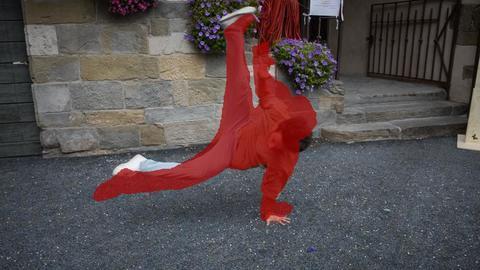} &
\includegraphics[width=0.16\textwidth,height=0.08\textheight]{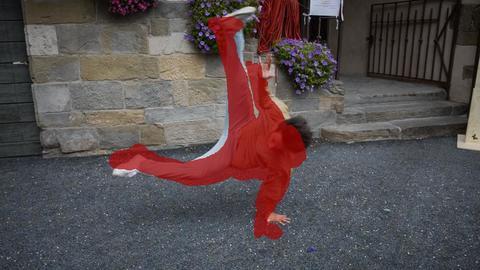} \\
Annotated image & \multicolumn{5}{c}{Example training masks}\\
\end{tabular}
\par\end{centering}
\caption{\label{fig:offline-augmentation2}Examples of training mask generation.
From one annotated image, multiple training masks are generated. The generated masks mimic plausible object shapes on the preceding frame.
}
\vspace{-1em}
\end{figure*}

\section{Dataset specific tuning}
\label{sec:dat_tun}
As mentioned in \S\ref{sec:single-frame-results} in the main paper by adding additional ingredients specifically tuned for
different datasets, such as optical flow and CRF post-processing, we can push the results even further, reaching
80.3 mIoU on DAVIS, 72.6 on YoutubeObjects and 70.3 on SegTrackv-2.
In this section we discuss the dataset specific tuning.

\paragraph{Optical flow}
Although optical flow can provide interesting gains, we
found it to be brittle when going across different datasets.
Therefore we explored different strategies to handle optical flow.

As discussed in \S\ref{sec:method-variants} of the main paper
given a video sequence, we compute the optical flow using EpicFlow \cite{EpicFlowCVPR15} with Flow Fields matches \cite{FlowFields15} and convolutional boundaries \cite{COB_Maninis16}.
In parallel to the $\mathtt{MaskTrack}$ with RGB images, we proceed to compute a second output mask using the magnitude of the optical flow field as input image (replicated into a three channel image).
We then fuse by averaging the output scores given by the two parallel networks (using RGB image and optical flow magnitude as inputs).

For DAVIS  we use the original $\mathtt{MaskTrack}$ model (trained with RGB images) as-is, without retraining. However, this strategy fails on YoutubeObjects and SegTrackv-2, mainly due to the failure modes of the optical flow
algorithm and its sensitivity to the video data quality. To overcome this limitation we additionally trained the $\mathtt{MaskTrack}$ model using optical flow magnitude images on video data instead of RGB images.
Training on optical flow magnitude images helps the network to be robust to the optical flow errors during the test time and provides a marginal improvement on YoutubeObjects and SegTrackv-2.

Overall integrating optical flow on top of $\mathtt{MaskTrack}$ provides $1\negmedspace\sim\negmedspace4 \%$ on each dataset.

\paragraph{CRF post-processing}
As have been shown in \cite{Chen2016ArxivDeeplabv2} adding on top a well-tuned post-processing CRF \cite{Kraehenbuehl2011Nips} can gain a
couple of mIoU points. Therefore following \cite{Chen2016ArxivDeeplabv2} we cross-validate the parameters of the fully connected CRF per each dataset based on the available first frame segment annotations of all video sequences.
We employ coarse-to-fine search scheme for tuning CRF parameters and fix the number of mean field iterations to 10. We apply the CRF on a temporal window of 3 frames to improve the temporal stability of the results. The color (RGB) and the spatio-temporal (XYT) standard deviation of the \emph{appearance kernel} are set, respectively, to 10 and 5. The pairwise term weight is set to 5. We employ an additional \emph{smoothness kernel} to remove small isolated regions. Both its weight and the spatial (XY) standard deviation are set to 1.

\section{Additional qualitative results}
\label{sec:add_qual_res}

In this section we provide additional qualitative results for the $\mathtt{MaskTrack}_{Box}$ and $\mathtt{MaskTrack}$ systems, described in \S\ref{sec:method} in the man paper.
Figure \ref{fig:qualitative-results2} shows the video object segmentation results when considering different types of annotations on DAVIS. Starting from segment annotations or even only from box annotations
on the first frame, our model generates high quality segmentations, making the system suitable for diverse applications.

\section{Examples of training mask generation}
\label{sec:mask_gen}

In \S\ref{sec:ablation-study} of the main paper we show that the main factor affecting the quality is using any form of mask deformations when creating the training samples (both for offline and online training).
The mask deformation ingredient is crucial for our $\mathtt{MaskTrack}$ approach, making the segmentation estimation more robust at test time to the noise in the input mask.
Figure \ref{fig:offline-augmentation2} complements Figure \ref{fig:offline-augmentation}  in the main paper and shows examples of generated masks using affine transformation as well as non-rigid deformations via thin-plate splines
(see \S\ref{sec:convnet-details} in the main paper for more details).

\section{Examples of optical flow magnitude images}
\label{sec:flow_gen}

\begin{figure*}
\begin{centering}
\begin{centering}
\begin{tabular}{@{}c@{  }c@{  }c@{  }c@{  }c@{}}

&\multicolumn{4}{c}{\bf DAVIS} \tabularnewline
\\
\begin{turn}{90}{\hspace{0.5em} RGB Image}
\end{turn}
&\includegraphics[width=0.24\textwidth,height=0.11\textheight]{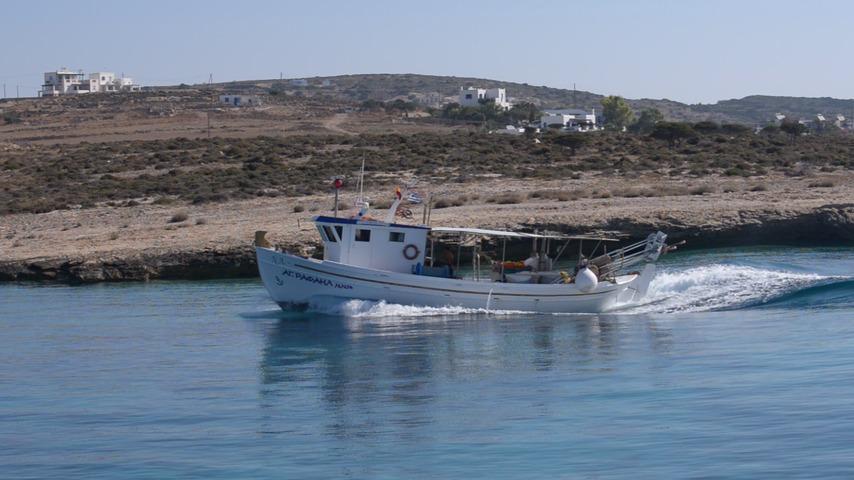} &
\includegraphics[width=0.24\textwidth,height=0.11\textheight] {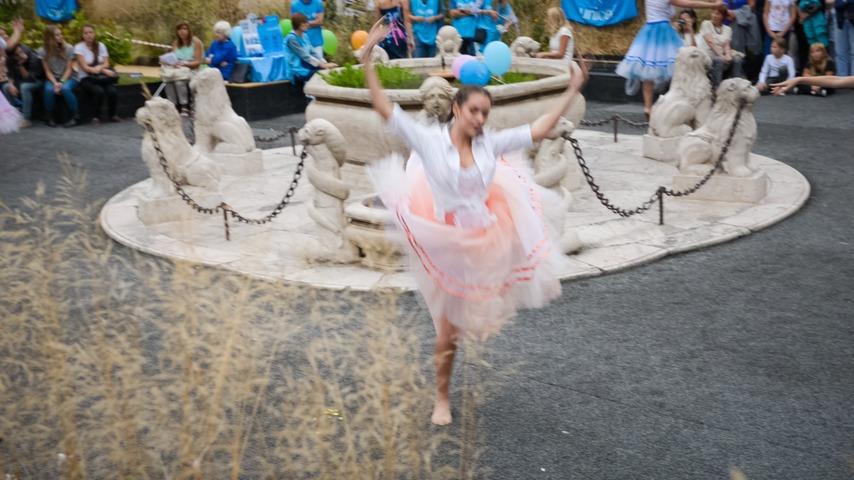} &
\includegraphics[width=0.24\textwidth,height=0.11\textheight] {figures/flow_images/image_horse_00005.jpg}&
\includegraphics[width=0.24\textwidth,height=0.11\textheight] {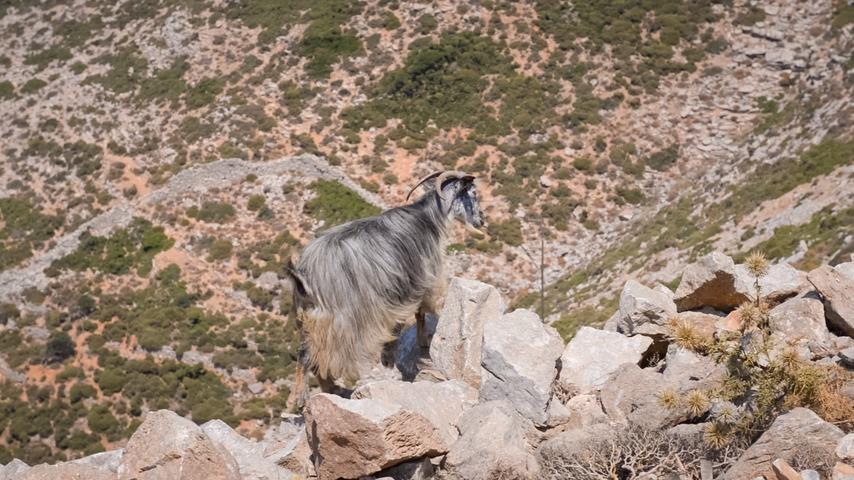} \tabularnewline
\begin{turn}{90}
{\hspace{0.5em} Optical Flow}
\end{turn}
&\includegraphics[width=0.24\textwidth,height=0.11\textheight]{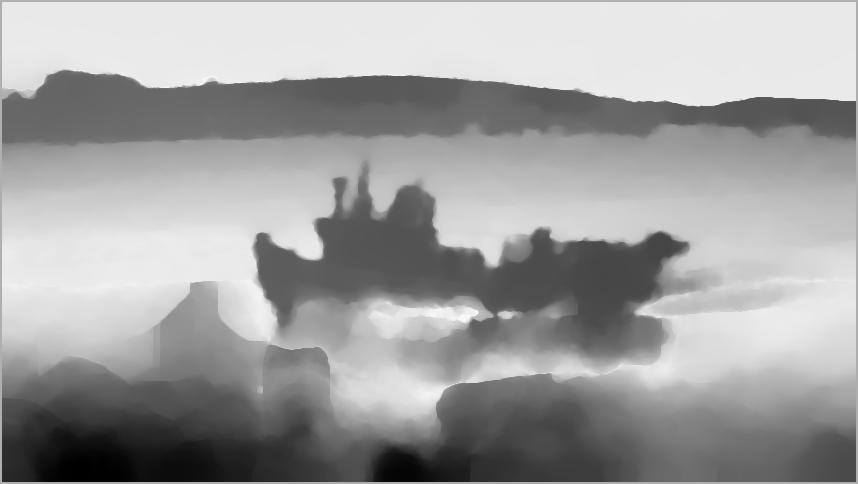} &
\includegraphics[width=0.24\textwidth,height=0.11\textheight] {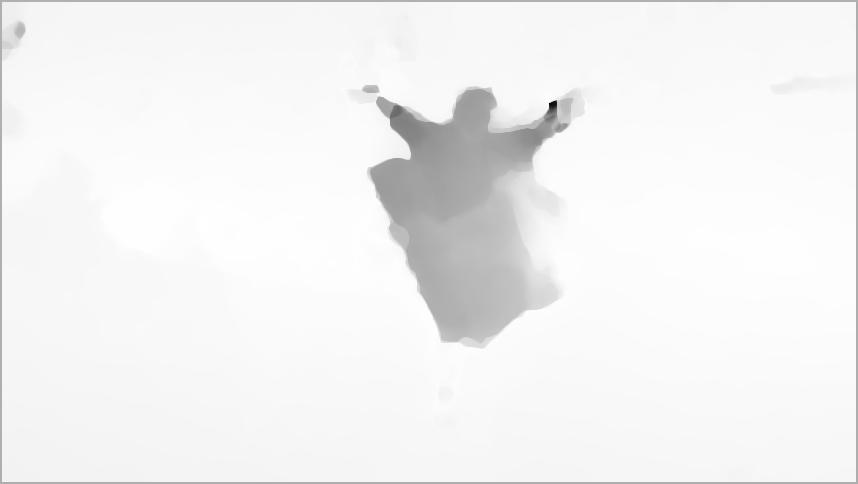} &
\includegraphics[width=0.24\textwidth,height=0.11\textheight] {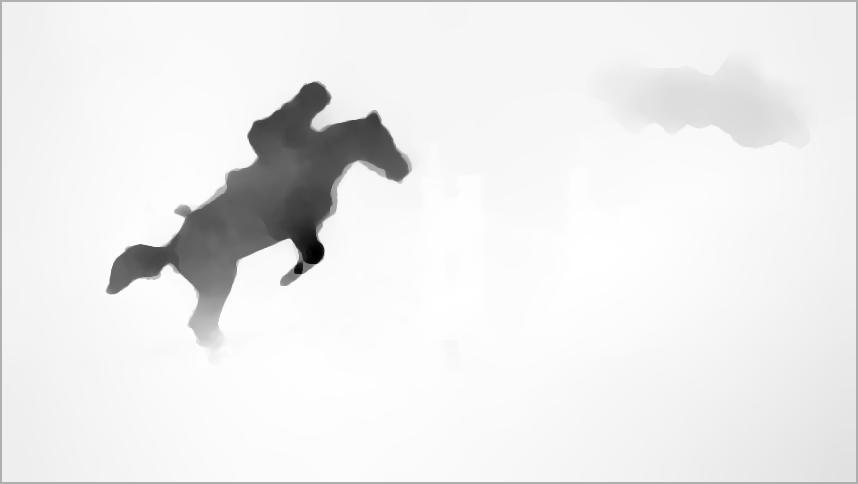}&
\includegraphics[width=0.24\textwidth,height=0.11\textheight] {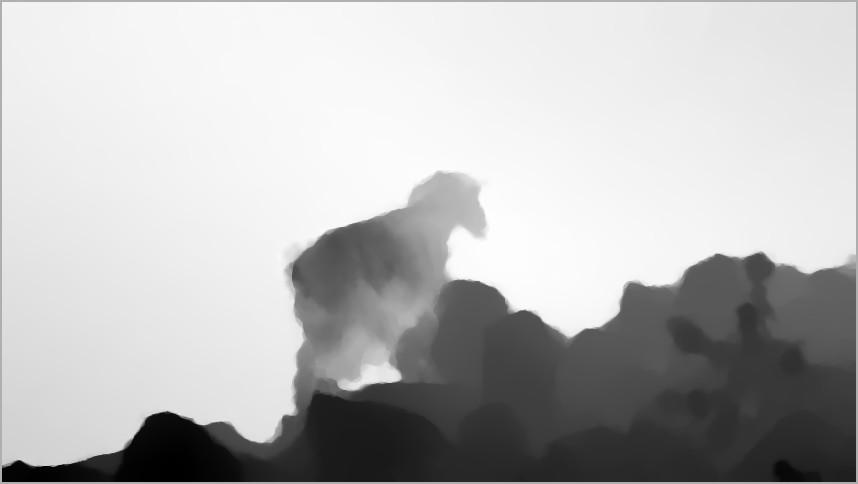} \tabularnewline
\\
\\
&\multicolumn{4}{c}{\bf SegTrack-v2} \tabularnewline
\\
\begin{turn}{90}{\hspace{0.5em} RGB Image}
\end{turn}
&\includegraphics[width=0.24\textwidth,height=0.11\textheight]{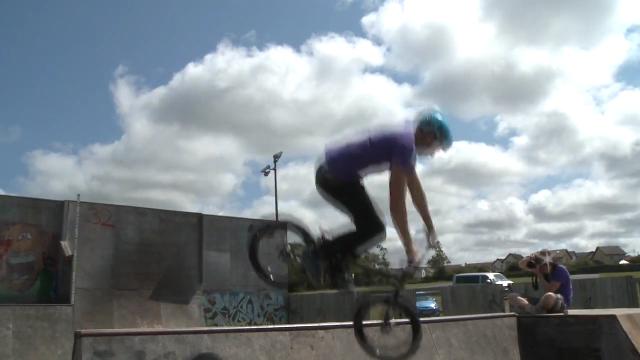} &
\includegraphics[width=0.24\textwidth,height=0.11\textheight] {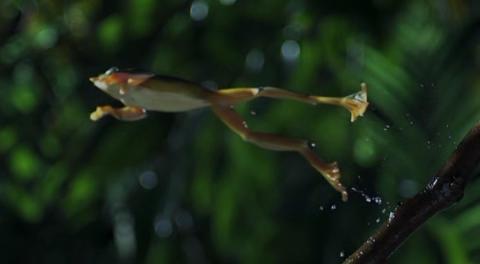} &
\includegraphics[width=0.24\textwidth,height=0.11\textheight] {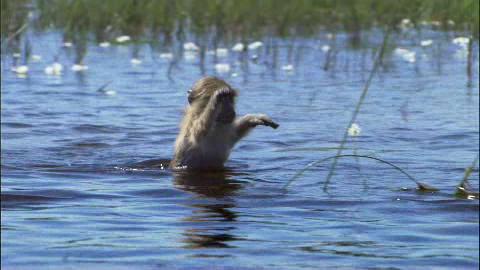}&
\includegraphics[width=0.24\textwidth,height=0.11\textheight] {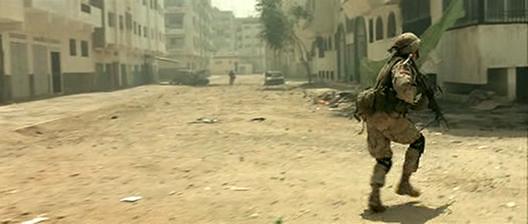} \tabularnewline
\begin{turn}{90}
{\hspace{0.5em} Optical Flow}
\end{turn}
&\includegraphics[width=0.24\textwidth,height=0.11\textheight]{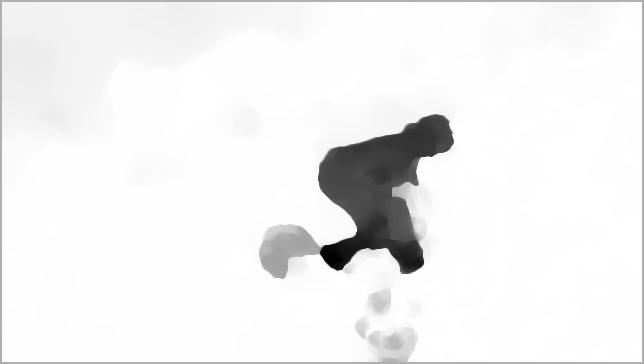} &
\includegraphics[width=0.24\textwidth,height=0.11\textheight] {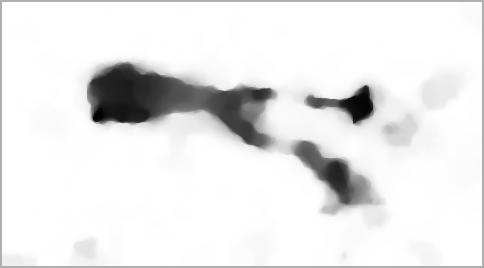} &
\includegraphics[width=0.24\textwidth,height=0.11\textheight] {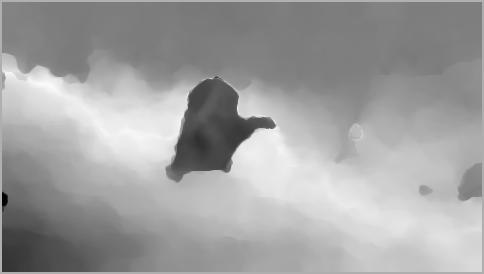}&
\includegraphics[width=0.24\textwidth,height=0.11\textheight] {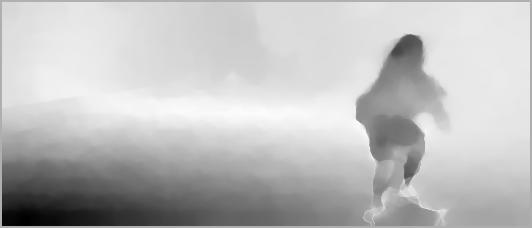} \tabularnewline
\end{tabular}
\par\end{centering}

\par\end{centering}
\caption{\label{fig:flow_images2}Examples of optical flow magnitude images for different datasets.}
\end{figure*}

In \S\ref{sec:method-variants} of the paper we propose to employ optical flow magnitude as a source of additional information to guide the segmentation.
The flow magnitude roughly looks like a gray-scale object and captures useful object shape information, therefore complementing the $\mathtt{MaskTrack}$ model with RGB images as inputs.
Examples of optical flow magnitude images are shown in Figure \ref{fig:flow_images2}. Figure \ref{fig:flow_images2} complements Figure \ref{fig:flow_images} in the main paper.

\end{document}